\PassOptionsToPackage{table,dvipsnames,xcdraw}{xcolor}
\documentclass[conference]{IEEEtran}

\usepackage{cite}
\usepackage{todonotes}

\usepackage[utf8]{inputenc} %
\usepackage[T1]{fontenc}    %
\usepackage{hyperref}       %
\usepackage{url}            %
\usepackage{booktabs}       %
\usepackage{nicefrac}       %
\usepackage{microtype}      %

\usepackage{wrapfig}

\usepackage{epsfig}
\usepackage{wrapfig}
\usepackage{pgfplotstable}
\usepackage{pgfplots}
\usepgfplotslibrary{groupplots}
\pgfplotsset{compat=newest} %
\usepgfplotslibrary{groupplots}
\usepgfplotslibrary{dateplot}
\usepackage{mathrsfs,amsmath,mathtools}
\usepackage{bbm}
\usepackage{tikz}
\usetikzlibrary{decorations.text,calc,shapes,arrows,arrows.meta, positioning,shapes.misc,decorations.markings,decorations.markings,decorations.pathreplacing,matrix,spy}

\usepackage{amssymb}
\usepackage{bbm}
\usepackage{dsfont}
\DeclareMathAlphabet{\mathmybb}{U}{bbold}{m}{n}
\newcommand{\1}{\mathmybb{1}}

\usepackage[noend]{algpseudocode}
\usepackage{algorithm}

\usepackage{rotating}

\pgfplotsset{
    discard if not/.style 2 args={
        x filter/.code={
            \edef\tempa{\thisrow{#1}}
            \edef\tempb{#2}
            \ifx\tempa\tempb
            \else
                
            \fi
        }
    }
}

\pgfplotscreateplotcyclelist{MyCyclelist}{%
  {Bittersweet, mark = *, dashed},
  {Blue, mark = square*, dashed},
  {Cyan, mark = diamond*, dashed},
  {DarkOrchid, mark = triangle*, dashed},
  {Green, mark = *, dashed},
  {Magenta, mark = square*, dashed},
  {Red, mark = diamond*, dashed},
  {Gray, mark = triangle*, dashed},
  {Black, mark = *, dashed},
    {Bittersweet, mark = square*, densely dotted},
  {Blue, mark = diamond*, densely dotted},
  {Cyan, mark = triangle*, densely dotted},
  {DarkOrchid, mark = *, densely dotted},
  {Green, mark = square*, densely dotted},
  {Magenta, mark = diamond*, densely dotted},
  {Red, mark = triangle*, densely dotted},
  {Gray, mark = *, densely dotted},
  {Black, mark = square*, densely dotted},
}

\usepackage{adjustbox}

\usepackage{booktabs}
\usepackage{multirow}

\usepackage{listings}
\title{Simpler Methods Work Better for L1 Penalized Logistic Models and Large Datasets}

\author{
\IEEEauthorblockN{Edward Raff}
\IEEEauthorblockA{
\textit{CrowdStrike, Inc.}, USA\\
edward.raff@crowdstrike.com\\
\textit{Univ. of Maryland, Baltimore County}, USA\\
raff.edward@umbc.edu
}
\and
\IEEEauthorblockN{James Holt}
\IEEEauthorblockA{
\textit{CrowdStrike, Inc.}, USA\\
james.holt1@crowdstrike.com
}
}

\begin{document}

\maketitle

\begin{abstract}
Linear models with an $L_1$-norm penalty remain state-of-the-art for high-dimensional (
$d > 1,000,000$) tasks, offering a straightforward method for solving real-world industry problems.
Despite their widespread use in industry and utility, many $L_1$ solvers are not effective for general use, are prohibitively slow, and are ineffective in parallelization. This makes them difficult
to train in an MLOps pipeline on large industry-scale corpora.
In this work, we test several proposed ``state-of-the-art'' solutions from the literature and find that older methods are currently far superior for general use. We also identify several recommendations for academics to perform research that avoids erroneously overconfident results, which can prevent the transition to production use.
Equally surprising, we find that a new and simple baseline, using LBFGS on a sub-gradient, is highly effective with minor tweaks, despite being dismissed in the literature for theoretical non-convergence. In practice, we find it is an easier-to-support and easier-to-scale method for production use.
\end{abstract}

\section{Introduction}

In this article, we are concerned with linear learning problems with a $d$-dimensional weight vector $\mathbf{w} \in \mathrm{R}^d$ learned over a dataset of $n$ samples $\mathbf{x}_i \in \mathbb{R}^d$ where it is often the case that $d \gtrapprox n$. In such cases, models that contain an $L_1$ penalty in the regularizer (also known as the LASSO~\cite{Tibshirani1994,Fu1998}) remain state-of-the-art in terms of classification or regression accuracy, primarily due to the challenge of learning in high-dimensional spaces and the robustness of the $L_1$-norm to irrelevant and spurious features~\cite{Ng2004}. The predominant generalized form of this problem is the Elastic-Net~\cite{Zou2005}, which balances a ratio of $\alpha$ between the $L_1$ and $L_2$ norm, for a given convex loss function $\ell(\cdot, \cdot)$, as shown in \autoref{eq:linear_l1_generic}. Our article will focus on the logistic loss, $\ell(x, y) = \log(1+\exp(-x \cdot y))$

\begin{equation} \label{eq:linear_l1_generic} f(\boldsymbol{w}) = 
    C \cdot \frac{1}{n}\sum_{i=1}^n \ell(\boldsymbol{x}_i^\top \boldsymbol{w}, y_i) + (1-\alpha) /2 \cdot \|\boldsymbol{w}\|_2^2 + \alpha \|\boldsymbol{w}\|_1 
\end{equation}

To be concise, we will use $f(\mathbf{w}) = \tilde{\ell}(\mathbf{w}) + \alpha \|\mathbf{w}\|_1$ to represent all components of the loss excluding the $L_1$ penalty ($\tilde{\ell}(\mathbf{w})$) separately. Another key note is that \autoref{eq:linear_l1_generic} is also often represented as $\tilde{\ell}(\mathbf{w}) + \lambda \alpha \|\mathbf{w}\|_1$, converting the parameter $C$ where $C \to \infty$ means less regularziation to $\lambda \to \infty$ means more regularization. 

We find that the research community has itself over-fit in evaluating progress in a manner inconsistent with how such models are used in practice. In short, \textit{the literature predominantly tests only the training loss on a small number of datasets, for far too strong a regularization strength $C$ that selects only $\approx 100$ features out of millions}. In regular use, much larger ranges of $C$/$\lambda$ strength are tested to better capture a sparsity (in solution $\mathbf{w}$) vs accuracy tradeoff. 

The primary contribution of our work is to show first (1) that this misalignment in evaluation vs. usage is that modern solver approaches are overly complicated and difficult to adapt and inordinately slow on realistic yet moderate penalties like $C=1$. 
The origin of this issue appears to be an over-reliance on computing the strongest penalty that yields $\mathbf{w}^*=\vec{0}$ as the starting penalty, and then scaling back from there. This penalty strength is calculated as $C_{min} = n/\|X^\top \boldsymbol{y}\|_\infty$ (or correspondingly, $\lambda_{max} = 1/(n C_{min})$). Many studies use a maximum value of $C = C_{min} \cdot 1000$, or a similar constant multiple, to test the ability to solve problems with ``many'' features. However, this is not realistic for many real-world use cases, and the value of $C$ is neither evaluated nor reported, nor is the number of features. Empirical evaluation shows that this is often too low a value to test, and that their optimizers fail to run on other reasonable datasets. 
Second (2), despite being dismissed as theoretically incapable due to the non-differentiability of $\|\mathbf{w}\|_1$ at the zero value, a lightly engineered application of the LBFGS optimizer is competitive with all existing approaches while being easier to adapt to new problems. However, attempting to ``engineer'' around the limitations of LBFGS proves to be easy and highly effective for large-scale classification problems.

The rest of this article is organized as follows. \autoref{sec:related_work} will review the related work and context of this article, and identify the key baselines we will evaluate in comparison. The seminal LBFGS algorithm will be used to implement a simple and strong baseline in \autoref{sec:baseline}. Our primary results will be evidenced in \autoref{sec:results} showing (a) how prior work considered too strong a penalty to be meaning, (b) that looking at the number of non-zeros, i.e. $\|\mathbf{w}\|_0$, would have helped avoid this issue, (c) that older and simpler coordinate descent approaches are still highly effective, and (d) that LBFGS with simple modifications is also very competitive --- and favorable even compared to versions modified to be theoretically sound with $L_1$ penalties. How LBFGS* enabled multiple use cases in a production environment, including large-, medium-, and small-scale model training, is discussed in \autoref{sec:production_use}. Finally, we will conclude in \autoref{sec:conclusion}.

\section{Related Work} \label{sec:related_work}

Our specific interest in this problem is the use of $ L_1$-penalized logistic regression in deploying malware classifiers over the past decade in industry and the U.S. federal government~\cite{malwareBytes}. Several thousand CPU cycles and research time have been devoted to improving and scaling these models due to their unique efficacy~\cite{lu_high-dimensional_2024,raff_saus,Raff2020autoyara,lu_optimizing_2026}: trained on n-grams of bytes, they produce small models ($\leq 2$ MB) that are highly deployable and run fast enough to filter and pre-screen data to avoid bogging down systems with excess work~\cite{raff_ngram_2016,Kilograms_2019,raff_zipf-gramming_2026,curtin_intermediate_2025}. The issue with $L_1$ models does not require malware data, which can't be shared due to various legal constraints~\cite{Harang2020,Joyce2022}, so we focus our experiments on publicly available data. However, we note that the largest dataset we consider in this article is 11 GB, and the issue is only more pronounced in our malware corpora, which can exceed 10 TB at small scales~\cite{joyce_ember2024_2025}. Indeed, the results are uninteresting at our production scale because only two methods run to completion; therefore, we present the (fiscally expensive) lessons learned from the pitfalls of modern $L_1$ works at an accessible and reproducible scale. First, we review the broad history and primary baselines of $L_1$ logistic regression, and then examine related work with respect to hyperparameter settings for $L_1$ regularization strength. Ultimately, our results reflect a long-standing trend in reproducibility and difficulties mapping what is needed by the practitioner, and what is described, being conflated with other factors~\cite{raff_what_2025,Raff2022a,raff_reproducibility_2023,Raff2022,Raff2020c,Raff2019_quantify_repro,Musgrave2020, coakley_examining_2022, Gundersen2018}. 

\subsection{Baselines and Broad History}

Many solvers for $L_1$ penalized problems have been proposed, and our work focuses on open-source and state-of-the-art solvers. Today, these are all based on coordinate descent. This approach was generally popularized by the seminal GLMNET solver~\cite{Friedman2010}, where one dimension $w_j$ is updated at a time and all $d$ dimensions are iterated until convergence. The improved ``NewGLMNET''~\cite{Yuan2012} poses one of our key benchmarks as an older, yet relatively simple and widely used via the LIBLINEAR library~\cite{Fan2008}.  

More recent solvers have incorporated concepts like screening rules~\cite{ElGhaoui2010,Tibshirani2012,wang_lasso_2013, Wang2014a, ndiaye_gap_2017, rakotomamonjy_screening_2019,dantas_expanding_2021,Larsson2021} to provably remove dimensions from the optimization, and Proxy-Newton steps~\cite{Lin2008}, in conjunction with accelerated gradient methods like Anderson's~\cite{bertrand_anderson_2021}, to compose new and sophisticated algorithms with many components. In particular, we will use \textsc{skglm} ~\cite{bertrand_beyond_2022} which purports to be ``huge scale'' and faster than other available methods, and merges many similar design choices from blitz~\cite{pmlr-v37-johnson15} and fireworks~\cite{rakotomamonjy_convergent_2022}. We also compare with \textsc{celer}, which is another working set, screening rule, coordinate descent-based solver with Acceleration -- but has two versions: one with proxy-Newton steps (\textsc{Celer-PN}) and one without (\textsc{Celer}), allowing us to ablate one component of the design. 

While coordinate descent has become the predominant approach in linear $L_1$ models, there is a smaller but non-trivial set of work that is row-based in using gradients with proximal operators or sub-gradients. The most widely used is likely the \textsc{SAGA} optimizer~\cite{NIPS2014_5258} that is the default $L_1$ solver in Scikit-Learn~\cite{scikit-learn}, and thus included as a benchmark. Another seminal work is that of \textsc{OWL-QN}~\cite{Andrew2007,Gong2015}, which we will include as a baseline for its flexibility and wide recognition. \textsc{OWL-QN} is also a key comparator as it is an extension of the L-BFGS algorithm to directly support $L_1$ penalized objectives, allowing us to compare to a similar and mathematically rigorous approach instead of our ``engineered'' solution. We will see that \textsc{OWL-QN} does not meaningfully improve upon our approach compared to other options, and is slower in many cases.   

While we will show a version of LBFGS that does work, most naive gradient methods like SGD, Adam, and batch gradients do not work well in practice~\cite{10882922}. There exists a variety of approaches that require doubling the number of features that have been proposed over time, which we avoid due to the high cost of doubling the number of features in high-dimensional problems~\cite{Zhou2015,ziyin_spred_2023,goldstein_split_2009,koh_method_2007,shi_fast_2010}. 
Works that are uniformly slower versions of already included methods are also excluded~\cite{dai_rehline_2023,blondel_lightning_2016,Friedman2010}.
This set includes approaches that re-cast $L_1$ problems as equivalent SVMs\cite{Zhou2015}, re-parameterizing the $L_1$ penalty as a combination of $L_2$ penalties~\cite{ziyin_spred_2023}, Bregman iterations~\cite{goldstein_split_2009}, and interior point methods~\cite{koh_method_2007}. The closest historical work is ~\cite{shi_fast_2010}, which proposed a hybrid approach: using gradient descent with line search to obtain a high-quality initial solution, followed by interior-point methods to refine it. In contrast, we do not consider any refinement step (e.g., apply coordinate descent afterward if desired if the specific values in $\mathbf{w}$ are important), and instead focus on improving the ``imprecise'' solver to be entirely sufficient on its own for obtaining sparse solutions with equivalent accuracy. 

\subsubsection{Other Libraries Not Presented}

We have discussed the primary baseline methods, which we will compare: newGLMNET, Celer, Celer-PN, SkGLM, and SAGA. These cover the current spectrum of just reviewed popular methods in academic state-of-the-art and most widely used libraries. We take a brief moment to also discuss some other libraries for linear learning that were evaluated, but will not be presented in the results for brevity and clarity.

First is the ReHLine library and method, which provides a framework for theoretically fast learning of linear models that can be composed into ReLU-like sub-components~\cite{dai_rehline_2023}. While this does not support the logistic loss, we did testing with an L1 penalized SVM and found that it exceeded our 24-hour limit in all tests. This issue appears because the result of ReHLine is all on datasets with $d<400$ dimensions and far more rows, and the L1 penalty is obtained by a re-parameterization that adds an additional $d$ rows to the training data. This is prohibitive in $n \leq d$ settings like we are interested in this work, and the primary utility domain of L1 models. The Lightning library ~\cite{blondel_lightning_2016} supports coordinate descent and SAGA, but is disadvantaged compared to our selected baselines in being a pure-python implementation and slower in our tests of moderately sized corpora. The seminal GLMNET is uniformly slower than newGLMNET (Liblinear) in all our tests, and so not duplicated~\cite{Friedman2010}.

\subsection{Hyper-Parameter Evaluations}

As we discussed the technical approaches in the literature, it is also convenient to discussed a pervasive issue in evaluation through the literature that is a key contribution of our work: the regularization penalty $\lambda$ is usually switch with respect to $\lambda_\mathit{max}$, the weakest penalty that results in the solution $\mathbf{w} =\vec{0}$. This corresponds to an equivalent $C_\mathit{min} = 1/(n \lambda_\mathit{max})$. 

Papers will focus only on $\lambda_{max}$ as the reference point, and will test smaller ratios of $\lambda/\lambda_{max}$. However, $\lambda_{max}$ is not ideal as a reference for penalties across datasets, because it is not normalized by the number of rows in the dataset $n$, where $C = 1/(\lambda n)$ by definition is normalized. Normalization by columns $d$ is relatively unnecessary as it is incorporated intrinsically into the penalty $\|\mathbf{w}\|_1$, and the purpose of an $L_1$ penalty is to be robust to uninformative/spurious additional features. This issue is widespread in the literature~\cite{massias_dual_2020,pmlr-v80-massias18a,barroso-luque_sparse-lm_2023,ren_thunder_2020,rakotomamonjy_convergent_2022,zeng_biglasso_2018,kim_another_2018}, and we will review it in more detail in our results. 

For example, \textsc{skglm} tested the largest dataset of all cited prior work, kdda, with a penalty ratio $\lambda/\lambda_{max} = 10^{-3}$. This corresponds to a penalty of $C=0.0001$, which is 100$\times$ smaller than our minimum value of $C=0.01$ and already yields a sparsity rate of 0.9998\%. While there are cases where extremely sparse solutions are the goal, 
it ignores that accuracy is often a major goal and usually obtained for $C \geq 1$ in the few works to evaluate a wide range of $C$ ~\cite{Fan2008,Yuan2012}.
The Celer papers \cite{massias_dual_2020,pmlr-v80-massias18a} consider only $\lambda_{\max}/20$ except when performing an experiment in comparing the number of features selected, at which point they find it necessary to evaluate $\log_{10}(\lambda_{max}/\lambda)=100$ which corresponds to $10^{100}$, a massive difference in the scale evaluated for the penalty. However, this was evaluated on a small dataset of only $n=72$ rows and $7,129$ features. The largest dataset in the Celer article is a regression dataset with $n = 16,087$ and $d = 1,668,738$ ($\approx$250MB in size), which benefits from closed-form updates under the mean-squared-error loss. 
Blitz uses the comparatively small RCV1 dataset with $\lambda_{max}/20$ and tests on the Webspam dataset, but only with $\lambda_{max}/100$ to ablate feature prioritization, without comparing with alternative solvers. 
The Fireworks paper used a subset of a larger dataset, without details on how the subset was chosen, and was limited to 2 million samples and small values of $\lambda$ corresponding to $C=0.0001$; it also did not test other large datasets \cite{rakotomamonjy_convergent_2022}. 
Thunder also tests only small datasets for overly sparse solutions, without considering accuracy~\cite{ren_thunder_2020}.
Some articles like Sparse-lm \cite{barroso-luque_sparse-lm_2023} have no tests, and simply report which kinds of norms and pseudo-norms it can find solutions for. 

The lack of attention to accuracy, the reliance on sparsity as a proxy for reasonableness, and the absence of testing on multiple large datasets are endemic issues in the current literature. Even recent results that aim to mitigate evaluation issues, such as benchopt~\cite{moreau_benchopt_2022}, still use only four small datasets and test with respect to $\lambda_\mathit{max}/100$, leaving the results non-informative for real-world solver needs and timing that test $\lambda$ by thousands of orders of magnitude. 
Our results show that the decade older Liblinear is still the fastest and most well-rounded solver.
Though $C$ is not the only reasonable criterion~\cite{Yang:2019:ENE:3292500.3330910}, the last work we found to extensively test $C$ or any alternative is over 10 years old~\cite{Yuan2012}, which may be another reason why the majority of solvers are insufficient for a large and important set of regular use.

\section{A Simple Baseline} \label{sec:baseline}

Research for training $L_1$ penalized models has continued to advance and become increasingly sophisticated. A key consideration in assessing whether such sophistication is necessary is whether simpler alternatives would have sufficed to produce the required solutions. In this section, we propose a simple baseline for training $L_1$-penalized classifiers using a lightly modified version of the seminal LBFGS algorithm, which we denote LBFGS*. LBFGS*'s purpose is to be an existence proof that other paths are worth exploring because they may yield faster convergence, lower runtime, or other advantages in computational efficiency --- and that the Pareto frontier of approaches to $L_1$ solving is not fully explored. 

The purpose of this approach is not to be better in all possible respects, and our approach will have the historically noted disqualifier: proving convergence is not possible due to the non-differentiability of $\|\mathbf{w}\|_1$ when any entry $w_j = 0$. This restriction seems obvious in that the purpose of using an $L_1$ penalty is to induce exact values of $w^*_j = 0$ in the solution $\mathbf{w}^*$ when solving \autoref{eq:linear_l1_generic}. As we will note in \autoref{sec:results}, this is not a significant issue in practice for large-scale classification problems. We make no claim or evaluation for other problems in which one may still desire an $L_1$ penalty, and the trade-offs our approach entails may not be sufficient. Our approach will make use of simple heuristics to improve convergence and runtime devised through experimentation: 
\begin{enumerate}
    \item The choice in gradient used will make a key difference in maintaining sparsity.
    \item The optimality criteria used in pruning/screening can be employed at minimal cost because it is already calculated ``for free''
    \item The non-convergence issues that do occur when applying just LBFGs can be largely mitigated via simple heuristics for using LBFGs with non-convex optimization problems, that also enhance runtime. 
\end{enumerate}

\subsection{Using the Minimum-Norm Subgradient as the ``Gradient''}

Our first choice concerns a known result that appears not to be applied to gradient-based baselines. 
Since $\|\mathbf{w}\|_1$ is non-differentiable at zero, what is technically required is a sub-gradient. If one uses a computer algebra system, automatic differentiation, or other tools, the standard solution returned will be to use the sub-gradient 
$\frac{\partial}{\partial \mathbf{w}} \|\mathbf{w}\|_1 = \operatorname{sgn}(\mathbf{w})$
, where $\operatorname{sgn}(\cdot)$ is the ``sign'' operator returning $\pm 1$ for positive/negative values respectively, and $0$ for any $0$ value.  This simple gradient can be seen in public implementations of LBFGS-B based approaches to the $L_1$ penalty\footnote{e.g., \url{https://gist.github.com/vene/fab06038a00309569cdd}}, usage in Tensorflow via the absolute value operator\footnote{e.g., \href{https://github.com/nfmcclure/tensorflow_cookbook/blob/master/03_Linear_Regression/04_Loss_Functions_in_Linear_Regressions/04_lin_reg_l1_vs_l2.py}{TensorFlow Cookbook example}}, and many other works that update all features simultaneously~\cite{liu_large-scale_2009,4407762,Figueiredo2003,Lawrence_etal_2004,Lawrence_etal_2005}. 
However, it is not necessary to consider the $L_1$ norm in isolation from its contribution to $\tilde{\ell}(\mathbf{w})$ to the objective function $f(\mathbf{w})$. To be specific, the subgradient $g$ must satisfy $f(\mathbf{w}) \geq f(\mathbf{z}) + g^\top (\mathbf{w}-\mathbf{z})$. If we add the constraint that $g$ is the minimum $\|g\|_2$ norm of all solutions (to make the solution unique), then we can derive the solution \autoref{eq:min_norm_subgradient}, where $[\cdot]^+ = \max(\cdot, 0)$. 
\begin{equation} \label{eq:min_norm_subgradient}
\nabla_j f(\boldsymbol{w}) \equiv 
\begin{cases}
\nabla_j \ell(\boldsymbol{w})+\alpha \operatorname{sgn}(\nabla_j \ell(\boldsymbol{w})) & \text { if } w_j \neq0 \\ 
\operatorname{sgn}\left(\nabla_j \ell(\boldsymbol{w})\right) \left[\left|\nabla_j \ell(\boldsymbol{w})\right|-\alpha\right]^+ & \text { if } w_j=0
\end{cases}
\end{equation}

This form is used in most coordinate descent works~\cite{massias_dual_2020,pmlr-v80-massias18a,barroso-luque_sparse-lm_2023,ren_thunder_2020,rakotomamonjy_convergent_2022,zeng_biglasso_2018,kim_another_2018} and is critical for maintaining sparsity as the solution $\mathbf{w}_t$ is initialized. Consider the value of the logistic loss, which is non-negative for all possible inputs. Thus, every gradient update will induce non-zero values in $\nabla \tilde{\ell}(\mathbf{w})$, and thus densify $\mathbf{w}$. By using the minimum norm subgradient, we maintain exact values of zero when necessary. 

\subsection{Pruning Irrelevant Dimensions} \label{sec:prunning}

One efficiency of using coordinate-descent-based solvers is that most dimensions have few nonzero values. Empirically, many datasets follow a power law to the frequency of feature use, and most informative features (i.e., retained in the solution of $\mathbf{w}^*$) tend to be frequent in occurrence, because features' occurrences are a prerequisite to making an informed prediction. A solution using only rare features is highly improbable, as it would require a large number of rare features to span the training samples. These rare features are removed from the optimization by pruning rules, thereby avoiding unnecessary work. 

For LBFGS to work on multiple dimensions simultaneously, this is potentially wasted work and entails a high computational cost. Consider the webspam dataset used in the next section, which has $n=350,000$ samples and $d=16,609,143$ features, and employs a history of $m=5$ in LBFGS. This means we will perform $d\cdot (2 m+1) = 182,700,573$ floating-point operations to compute the search direction. If only a small number of non-zeros will exist in the solution $\mathbf{w}$, this is a large amount of work that is performed irrespective of the solution sparsity. 

To mitigate this issue, we leverage the observation first reported in \cite{Yuan2010}, that at the solution  $w^*_{j} = 0$ if and only if $|\nabla_j \tilde{\ell}(\mathbf{w}^*)| < \alpha $, which corresponds to the subgradeint rule of \autoref{eq:min_norm_subgradient}.  \cite{Yuan2010} further proved that this will eventually hold for each iterate toward the solution, but which iterate will make this true for all future iterations is unknown. We instead employ a simple heuristic to check $|\nabla_j \tilde{\ell}(\mathbf{w}_t)| < \alpha \beta$ where $\beta = 1.0$ is the start of a line search, where $\beta$ is reduced until the objective $f(\mathbf{w}_t)$ is increased by at most a de minimis amount. In this manner we can adapt the optimization to the set of active parameters in a way that requires no extra mechanics (we are already calculating sub-gradients), and avoid excess work. This has the benefit of inducing sparsity in the solution by pruning non-zeros and maintaining it by removing variables from the optimization once they are set to zero. Because LBFGS does not guarantee convergence in theory, we allow the search space to expand or contract iteratively. 

\subsection{A Full Baseline Methods}

Our approach to using the minimum norm sub-gradient in \autoref{eq:min_norm_subgradient} and running multiple dimensions are combined with a number of heuristics to make LBFGS reliable for solving large-scale $L_1$ logistic regression. The entire procedure is summarized in \autoref{alg:brent-lbfgs}, where we discuss the modifications we have made to improve reliability. The discussion will proceed sequentially through the pseudo-code

\begin{algorithm}[!h]
\caption{Gradient--Brent--LBFGS Optimization with Armijo Line Search}
\label{alg:brent-lbfgs}
\begin{algorithmic}[1]
\Require Objective $f:\mathbb{R}^d\!\to\!\mathbb{R}$, tolerances $f_{\text{tol}},\,g_{\text{tol}}$
\State $\mathbf{w}_{-1} \gets  -\nabla f(\vec{\mathbf{0}})$
\State $\mathbf{w}_{0} \gets  \eta \cdot \mathbf{w}_{-1}$, s.t. $\eta = \underset{\eta^*}{\operatorname{arg min}}  f(\eta^* \cdot \mathbf{w}_{-1}) $ \Comment{{\color{ForestGreen}1-D optimization }}
\State $\mathsf{EMA}_0 \gets \infty$, \quad $t\gets 0$, \quad history $\mathcal{H}\gets\varnothing$
\State $\gamma \gets 10^{-2}$
\While{$\mathsf{EMA}_t > f_{\text{tol}}$ \textbf{and} $\|\nabla f(\mathbf{w}_t)\|_{\infty} > g_{\text{tol}}$}
    \State $\mathbf{p}_t \gets -\textsc{LBFGS\_Direction}\!\bigl(\nabla f(\mathbf{w}_t),\mathcal{H}\bigr)$
    \State $\alpha_t \gets \textsc{ArmijoLineSearch}\!\bigl(f,\mathbf{w}_t,\mathbf{p}_t\bigr)$
    \If{$\alpha_t \leq 0.001$}                     \Comment{\emph{\color{ForestGreen} Reset on failed search}}
        \If{$\mathcal{H} = \varnothing$}
            \State \Return $\mathbf{w}_t$ \Comment{{\color{ForestGreen}Converged in small steps}}
        \EndIf
        \State $\mathcal{H}\gets\varnothing$            \Comment{{\color{ForestGreen}clear $\{(\mathbf{s},\mathbf{y})\}$ pairs}}
        \State \textbf{continue}                       \Comment{{\color{ForestGreen}retry from current $\mathbf{w}_t$}}
    \EndIf
    \State $\mathbf{w}_{t+1} \gets \mathbf{w}_t + \alpha_t\,\mathbf{p}_t$
    \State $\Delta f_{t+1} \gets \lvert f(\mathbf{w}_{t+1})-f(\mathbf{w}_t)\rvert$
    \State $\mathsf{EMA}_{t+1} \gets 0.1\,\Delta f_{t+1} + (1-0.1)\,\mathsf{EMA}_t$
    \State $\mathbf{s}_t \gets \mathbf{w}_{t+1}-\mathbf{w}_t,\quad
           \mathbf{y}_t \gets \nabla f(\mathbf{w}_{t+1})-\nabla f(\mathbf{w}_t)$
    \State $\mathcal{H}\gets (\mathcal{H}\cup\{(\mathbf{s}_t,\mathbf{y}_t)\})$  \Comment{{\color{ForestGreen}keep latest $m$ pairs}}
    \If{$\mathsf{EMA}_{t+1}< \gamma$} \Comment{{\color{ForestGreen}Check if we should eliminate variables into the optimization}}
        \State $\beta \gets 1$, \quad $\mathcal{A} \gets \1 \left[|\nabla \tilde{\ell}(\mathbf{w})| \geq \beta\right]$
        \While{$f\left(\mathbf{w}\odot \mathcal{A}\right) > f(\mathbf{w}) \cdot 1.001$}
            \State $\beta \gets \beta/2$,\quad $ \mathcal{A} \gets \1 \left[|\nabla \tilde{\ell}(\mathbf{w})| \geq \beta\right]$
        \EndWhile
        \State $\mathbf{w}_{t+1} \gets $ active values of $\mathcal{A}$ \Comment{{\color{ForestGreen}We alter the dimension of the problem}}
        \State $\mathcal{H} \gets \varnothing$ \Comment{{\color{ForestGreen}Old gradient history has the wrong size}}
        \State $\gamma \gets \mathsf{EMA}_{t+1}/10$ \Comment{{\color{ForestGreen}Check for active features in another order of magnitude reductions}}
    \EndIf
    \State $t \gets t+1$
\EndWhile
\State \Return $\mathbf{w}_t$
\end{algorithmic}
\end{algorithm}

First, it is necessary to select the initial solution $\mathbf{w}_0$. It is common to use the zero vector $\vec{0}$ as the initial value. When computing a regularization path, sequential values of $C$ are evaluated and used as the initial value for the next optimization; however, this is beyond the scope of the current study. Our first insight is that, due to the $L_1$ penalty, the initial weights are often not properly scaled relative to the solution, causing a large number of updates to the gradient that primarily change magnitude rather than the solution. To mitigate this, line 2 begins with a one-dimensional optimization problem, where we find a scalar $\eta$ such that the gradient of the zero-vector solution is minimized with respect to the loss function. 
We use Brent's ~\cite{Brent1971}, available in Scipy,
to solve this optimization which can be solved efficiently by computing $\hat{z} = X \mathbf{w}$, and noting that $\eta \hat{z} = X \eta \mathbf{w}$ and $\|\mathbf{w} \hat{z}\|_1 = \hat{z} \|\mathbf{w} \|_1$. Thus $f(\mathbf{w} \hat{z})$ can be evaluated any number of times for just one matrix-vector product, making it computationally faster than using any gradient information directly. 

We use the common convergence criterion that the $L_\infty$ norm of the gradient is below a tolerance $g_\mathit{tol}$ (0.001 in our experiments). However, in some cases, we find that the solution is effectively converged long before this criterion is hit. So our second improvement on line 3 (and later 5 and 15) is to use an exponential moving average (EMA) of the difference in function values $\Delta f_{t+1} \gets \lvert f(\mathbf{w}_{t+1})-f(\mathbf{w}_t)\rvert$ to detect convergence. While the use of an $f_\mathit{tol}$ is common, it is noisy in that a single iteration may have a small difference and cause a premature termination. Using the EMA mitigates this issue; we assign a weight of 0.1 to the most recent iterate without any tuning. 

Using $\mathcal{H}$ to represent the running $m$ most recent items in the LBFGS memory, our preference is to use the Armijo line search as it will return a value of $\alpha_t = 1$ for most iterations (by design~\cite{Armijo1966}) and require no extra evaluations. However, as noted in the literature, LBFGS can fail its line search due to the non-differentiability of $\|\mathbf{w}\|_1$. Our improved initialization mitigates much of this problem, and we further remediate it by clearing the history whenever an apparently too-small line search occurs on lines 8-12. Together, we find this sufficient for all seven datasets we consider, and it continues to perform well on many smaller datasets in extended testing. 

Finally, lines 18-25 incorporate the line search to prune unused dimensions as discussed in \autoref{sec:prunning}, where $\mathcal{A}$ represents a binary mask of the ``active'' dimensions. This check is applied for every order-of-magnitude decrease in the EMA of $\Delta f_{t+1}$ to avoid excessive computation, since it requires creating a new (reduced) copy of $X$ in memory, which is moderately expensive. However, on many datasets, this allows for subsequent iterations to run much faster.

\section{$L_1$ Results} \label{sec:results}

We emphasize that the purpose of our method is not to be superior to all other methods for solving $L_1$ constrained problems. This work shows that the literature has focused too heavily on an evaluation criterion that does not match real-world needs. Part of demonstrating this is showing that the current ``state-of-the-art'' methods do not perform as well as prior approaches in coordinated descent-based optimization. Similarly, our LBFGS* approach from \autoref{sec:baseline} is composed entirely of older techniques, yet often outperforms other approaches. This, in conjunction, shows that there is likely a dearth of consideration for alternative approaches. 

At a minimum, the desirata for a general-purpose approach to solving $L_1$ penalties should: 
    (1) Have reasonable runtimes over the range of regularly tested $C$ penalty values, which has been at least the range [1/100, 100] for multiple decades~\cite{Chang2011,Platt1998,Lawrence_etal_2005,koh_method_2007,Yuan2012}.
    (2) Have a reasonable runtime for the value of $C$ that empirically produces the best accuracy on a validation set. 
    (3) Unless the dataset is otherwise degenerate (i.e., collinear), the method should be usable for cold-start penalties that select at least 10\% of the $d$ total features. 

The first concerns the practical relevance for practitioners of common libraries, as users may understand nothing beyond the documentation and tutorials indicating that $L_1$ is suitable for sparse solutions~\cite{scikit-learn}. The second is a matter of utility in cases where maximal accuracy is desired, and the third condition is the converse, where it is desired to explore a spectrum of sparse solutions. By being able to cold-start (i.e., initial weight $\mathbf{w}_0 = \vec{0}$) a solution under each condition, a method should be at least minimally viable for common use.

Our first insight into this issue stems from testing multiple larger datasets, when most articles only evaluate zero or one dataset of 1 GB in size. These datasets are summarized in \autoref{tbl:dataset_sizes}, showing the value of $C_\mathit{min}$ that produces the zero-vector solution $\mathbf{w}^*=0$. Because prior work has mostly evaluated with respect to $C_\mathit{min}/C$ (or equivalently, $\lambda/\lambda_\mathit{max}$), the exact value of the penalty is never expressed. Since they test at most $C = C_\mathit{min} \cdot 1000$, \textit{they do not expose or realize that the entire testing procedure is done over extreme regularization penalties that produce $\leq 100$ non-zero vectors in the solution}. This is because the scaling in the number of non-zeros $\mathbf{w}$ has no pre-defined relationship with the scaling of the penalty $C$, 

\begin{table}[!h]
    \centering
\caption{Most papers consider only one, or no, datasets of these sizes. Yet these datasets are imminently reasonable to run on a modern desktop, and the minimum value of $C$ (right column) that produces $\mathbf{w}^*=0$ is always at least 5 orders of magnitude smaller than $C=1$, yet it is a common default. }
\label{tbl:dataset_sizes}
\adjustbox{max width=0.9\columnwidth}{%
\begin{tabular}{@{}lrrrc@{}}
\toprule
\multicolumn{1}{c}{Dataset} & \multicolumn{1}{c}{$n$} & \multicolumn{1}{c}{$d$} & \multicolumn{1}{c}{GBs} & $C_{min}$  \\ \midrule
avazu-app                   & 12,642,186              & 1,000,000               & 1.57                    & $4.42 \cdot 10^{-07}$ \\
avazu-site                  & 23,567,843              & 1,000,000               & 2.92                    & $2.83 \cdot 10^{-07}$ \\
url                         & 2,396,130               & 3,231,961               & 2.23                    & $9.00 \cdot 10^{-06}$ \\
kdda                        & 8,407,752               & 20,216,830              & 2.48                    & $1.07 \cdot 10^{-06}$ \\
kddb                        & 19,264,097              & 29,890,095              & 4.61                    & $4.32 \cdot 10^{-07}$ \\
kdd12                       & 149,639,105             & 54,686,452              & 11.01                   & $3.52 \cdot 10^{-08}$ \\
webspam                     & 350,000                 & 16,609,143              & 10.44                   & $4.55 \cdot 10^{-05}$ \\ 
\bottomrule
\end{tabular}
}
\end{table}

\textbf{Recommendation:} Evaluating across a sequence of solution sparsities/regularization strengths is sensible, but expressing that relationship as $\lambda/\lambda_\mathit{max}$ risks inadvertently obscuring the range of meaningful values. Instead, a range of $ C\in [10^{-2}, 10^2]$ should be considered, or plotting with respect to a range of non-zeros ~\cite{koh_method_2007,Tibshirani2012} in the solution (which can be linearly or logarithmically spaced as appropriate to the problem at hand) avoids obscuring vital details in the result.

Next, we examine each method's performance using only two penalties, as these suffice to illustrate the issue while keeping computation reasonable. First, we will evaluate $C=0.01$, which is already larger than the values considered in our baseline methods, Skglm, Celer, and SAGA. Second, we will look at $C=1.0$, the default value in Scikit-Learn~\cite{scikit-learn} and indeed most other libraries~\cite{Abeel2009}. The results are shown in \autoref{tbl:main_results}, where each solver is run to the original papers' specification: a $g_\mathit{tol} = 10^{-4}$ for Skglm, Celer (in appendix for being uniformly worse than Celer-PN), Celer-PN, and SAGA, $10^{-2}$ for Liblinear, and $10^{-3}$ for OWL-QN and LBFGS*. Comparing methods by solver tolerance is challenging beacuse each method uses slightly different stopping conditions. We bias the results in favor of Skglm, Celer, Celer-PN, and SAGA, by testing sequentially increasing tolerances from $10^{-1}$, $10^{-1}/5$, $10^{-2}$, $\ldots$, $10^{-4}$ whenever convergence was not achieved at their original tolerance criteria. {\color[HTML]{CB0000} Entries in red show cases where the solver at the target tolerance could not converge within 24 hours, and are excluded from any ``best'' result due to unintended early termination.}  This shows that even significantly constraining in favor of prior methods, \textit{they are failing to converge for any meaningful tolerance threshold}. 

\begin{table}[!h]
\caption{Results showing the training loss function $f(\boldsymbol{w})$, test accuracy (Acc), number of non-zeros in the solution (nnz), and training time in Seconds. Best results are marked in \textbf{bold}. {\color[HTML]{CB0000} Entries in red indicate models that failed to converge, or could not converge within 24 hours of runtime, and were excluded for ``best'' designations due to failure}. All timing results are from an Apple M3 with 12 cores and 96 GB of RAM.}\label{tbl:main_results}
\centering
\adjustbox{max width=0.9\columnwidth}{%
\begin{tabular}{llrrrrrr}
\hline
\multicolumn{2}{c}{Dataset}                                                          & \multicolumn{1}{c}{skglm}     & \multicolumn{1}{c}{Celer-PN}   & \multicolumn{1}{c}{SAGA}      & \multicolumn{1}{c}{Liblinear} & \multicolumn{1}{c}{OWL-QN}    & \multicolumn{1}{c}{LBFGS*} \\ \hline
                                                                              & f(w) & 0.2065                        & 0.2065                         & \textbf{0.2064}               & 0.3346                        & 0.3259                        & 0.3571                    \\
                                                                              & Acc  & 93.5                          & 93.6                           & 93.5                          & \textbf{93.7}                 & 93.5                          & 93.1                      \\
                                                                              & nnz  & 50                            & 50                             & 55                            & 105                           & 53                            & 398                       \\
\multirow{-4}{*}{\begin{tabular}[c]{@{}l@{}}gisette\\ C=0.01\end{tabular}}    & Time & 0.8                           & 3.6                            & 11.5                          & \textbf{0.4}                  & 48.5                          & 16.3                      \\ \hline
                                                                              & f(w) & 0.0062                        & \textbf{0.0061}                & 0.0616                        & 0.0310                        & 0.0251                        & 0.0305                    \\
                                                                              & Acc  & 98.0                          & 98.1                           & 97.9                          & 97.9                          & \textbf{98.2}                 & 98.0                      \\
                                                                              & nnz  & 699                           & 631                            & 2392                          & 2144                          & 1098                          & 3410                      \\
\multirow{-4}{*}{\begin{tabular}[c]{@{}l@{}}gisette\\ C=1\end{tabular}}       & Time & 6.5                           & 817.2                          & 15.6                          & \textbf{0.8}                  & 221.0                         & 10.1                      \\ \hline
                                                                              & f(w) & 0.0905                        & \textbf{0.0904}                & 0.0911                        & 0.0986                        & 0.1228                        & 0.1217                    \\
                                                                              & Acc  & 97.0                          & \textbf{97.1}                  & 97.0                          & 96.8                          & 96.8                          & 96.8                      \\
                                                                              & nnz  & 79                            & 87                             & 267                           & 155                           & 246                           & 515                       \\
\multirow{-4}{*}{\begin{tabular}[c]{@{}l@{}}URL\\ C=0.01\end{tabular}}        & Time & 33.2                          & 327.3                          & 219.9                         & \textbf{6.7}                  & 60.6                          & 84.5                      \\ \hline
                                                                              & f(w) & 0.0598                        & {\color[HTML]{CB0000} 0.0381}  & \textbf{0.0396}               & 0.0571                        & 0.0740                        & 0.0528                    \\
                                                                              & Acc  & 97.9                          & {\color[HTML]{CB0000} 98.7}    & \textbf{98.6}                 & 98.3                          & 97.5                          & 98.4                      \\
                                                                              & nnz  & 471                           & {\color[HTML]{CB0000} 1681}    & 2391                          & 3056                          & 13925                         & 11843                     \\
\multirow{-4}{*}{\begin{tabular}[c]{@{}l@{}}URL\\ C=1\end{tabular}}           & Time & 31.5                          & {\color[HTML]{CB0000} 8832.7}  & 1268.8                        & \textbf{8.7}                  & 42.4                          & 89.8                      \\ \hline
                                                                              & f(w) & \textbf{0.2224}               & 0.2225                         & 0.2227                        & 0.2227                        & 0.3050                        & 0.3095                    \\
                                                                              & Acc  & 92.5                          & 92.5                           & 92.5                          & 92.5                          & 92.5                          & \textbf{92.7}             \\
                                                                              & nnz  & 48                            & 47                             & 48                            & 50                            & 68                            & 177                       \\
\multirow{-4}{*}{\begin{tabular}[c]{@{}l@{}}Webspam\\ C=0.01\end{tabular}}    & Time & \textbf{18.8}                 & 48.1                           & 520.4                         & 28.2                          & 559.5                         & 448.2                     \\ \hline
                                                                              & f(w) & 0.0417                        &                                &                               & \textbf{0.0383}               & 0.0674                        & 0.0609                    \\
                                                                              & Acc  & 98.6                          &                                &                               & \textbf{99.0}                 & 98.4                          & 98.8                      \\
                                                                              & nnz  & 486                           &                                &                               & 1597                          & 6070                          & 6711                      \\
\multirow{-4}{*}{\begin{tabular}[c]{@{}l@{}}Webspam\\ C=1\end{tabular}}       & Time & 24.0                          & \textgreater{}24 hours         & \textgreater{}24 hours        & \textbf{23.6}                 & 219.5                         & 310.4                     \\ \hline
                                                                              & f(w) & 0.3099                        & \textbf{0.3093}                & {\color[HTML]{CB0000} 0.2989} & 0.3158                        & 0.3150                        & 0.3174                    \\
                                                                              & Acc  & 99.8                          & 99.7                           & {\color[HTML]{CB0000} 99.2}   & 99.7                          & 99.7                          & \textbf{99.9}             \\
                                                                              & nnz  & 551                           & 609                            & {\color[HTML]{CB0000} 10188}  & 663                           & 657                           & 1356                      \\
\multirow{-4}{*}{\begin{tabular}[c]{@{}l@{}}Avazu-app\\ C=0.01\end{tabular}}  & Time & 72.7                          & 6236.1                         & {\color[HTML]{CB0000} 3042.1} & 35.7                          & 195.8                         & \textbf{30.0}             \\ \hline
                                                                              & f(w) & {\color[HTML]{CB0000} 0.3214} & {\color[HTML]{CB0000} 0.2988}  & {\color[HTML]{CB0000} 0.2983} & \textbf{0.3006}               & {\color[HTML]{CB0000} 0.3769} & 0.3065                    \\
                                                                              & Acc  & {\color[HTML]{CB0000} 99.8}   & {\color[HTML]{CB0000} 99.2}    & {\color[HTML]{CB0000} 99.2}   & 98.3                          & {\color[HTML]{CB0000} 84.0}   & \textbf{99.5}             \\
                                                                              & nnz  & {\color[HTML]{CB0000} 115}    & {\color[HTML]{CB0000} 9597}    & {\color[HTML]{CB0000} 22928}  & 10209                         & {\color[HTML]{CB0000} 26}     & 10402                     \\
\multirow{-4}{*}{\begin{tabular}[c]{@{}l@{}}Avazu-app\\ C=1\end{tabular}}     & Time & {\color[HTML]{CB0000} 34.0}   & {\color[HTML]{CB0000} 31768.6} & {\color[HTML]{CB0000} 9408.6} & \textbf{20.5}                 & {\color[HTML]{CB0000} 0.3}    & 47.1                      \\ \hline
                                                                              & f(w) & \textbf{0.4424}               & {\color[HTML]{CB0000} 0.4416}  &                               & 0.4462                        & 0.4456                        & 0.4516                    \\
                                                                              & Acc  & 80.7                          & {\color[HTML]{CB0000} 98.0}    &                               & 98.0                          & 98.0                          & \textbf{98.3}             \\
                                                                              & nnz  & 814                           & {\color[HTML]{CB0000} 922}     &                               & 989                           & 1053                          & 3692                      \\
\multirow{-4}{*}{\begin{tabular}[c]{@{}l@{}}Avazu-site\\ C=0.01\end{tabular}} & Time & 148.8                         & {\color[HTML]{CB0000} 16475.8} & \textgreater{}24 hours        & 75.1                          & 416.5                         & \textbf{34.8}             \\ \hline
                                                                              & f(w) & 0.4412                        &                                &                               & \textbf{0.4369}               & 0.4367                        & 0.4446                    \\
                                                                              & Acc  & \textbf{98.0}                 &                                &                               & 97.7                          & 97.7                          & \textbf{98.0}             \\
                                                                              & nnz  & 1007                          &                                &                               & 11798                         & 12186                         & 11200                     \\
\multirow{-4}{*}{\begin{tabular}[c]{@{}l@{}}Avazu-site\\ C=1\end{tabular}}    & Time & 114.6                         & \textgreater{}24 hours         & \textgreater{}24 hours        & \textbf{42.8}                 & 296.7                         & 54.9                      \\ \hline
                                                                              & f(w) & \textbf{0.3180}               &                                &                               & 0.3326                        & 0.3324                        & 0.3347                    \\
                                                                              & Acc  & \textbf{88.8}                 &                                &                               & 88.7                          & \textbf{88.8}                 & 88.5                      \\
                                                                              & nnz  & 3845                          &                                &                               & 4002                          & 4394                          & 11130                     \\
\multirow{-4}{*}{\begin{tabular}[c]{@{}l@{}}kdda\\ C=0.01\end{tabular}}       & Time & 182.5                         & \textgreater{}24 hours         & \textgreater{}24 hours        & \textbf{41.5}                 & 725.5                         & 71.5                      \\ \hline
                                                                              & f(w) & {\color[HTML]{CB0000} 0.3222} &                                &                               & 0.2668                        & \textbf{0.2648}               & 0.2855                    \\
                                                                              & Acc  & {\color[HTML]{CB0000} 88.5}   &                                &                               & 89.3                          & \textbf{90.4}                 & 89.5                      \\
                                                                              & nnz  & {\color[HTML]{CB0000} 3771}   &                                &                               & 760537                        & 958627                        & 1576491                   \\
\multirow{-4}{*}{\begin{tabular}[c]{@{}l@{}}kdda\\ C=1\end{tabular}}          & Time & {\color[HTML]{CB0000} 59.8}   & \textgreater{}24 hours         & \textgreater{}24 hours        & \textbf{65.8}                 & 869.7                         & 218.1                     \\ \hline
                                                                              & f(w) & \textbf{0.3143}               &                                &                               & 0.3263                        & 0.3256                        & 0.3278                    \\
                                                                              & Acc  & \textbf{89.1}                 &                                &                               & \textbf{89.1}                 & \textbf{89.1}                 & \textbf{89.1}             \\
                                                                              & nnz  & 8361                          &                                &                               & 8509                          & 9533                          & 22932                     \\
\multirow{-4}{*}{\begin{tabular}[c]{@{}l@{}}kddb\\ C=0.01\end{tabular}}       & Time & 159.0                         & \textgreater{}24 hours         & \textgreater{}24 hours        & \textbf{72.0}                 & 1048.4                        & 176.4                     \\ \hline
                                                                              & f(w) & {\color[HTML]{CB0000} 0.3291} &                                &                               & 0.2523                        & \textbf{0.2520}               & 0.2814                    \\
                                                                              & Acc  & {\color[HTML]{CB0000} 88.9}   &                                &                               & 89.8                          & \textbf{89.9}                 & 89.8                      \\
                                                                              & nnz  & {\color[HTML]{CB0000} 2711}   &                                &                               & 1775533                       & 2202695                       & 2983383                   \\
\multirow{-4}{*}{\begin{tabular}[c]{@{}l@{}}kddb\\ C=1\end{tabular}}          & Time & {\color[HTML]{CB0000} 47.6}   & \textgreater{}24 hours         & \textgreater{}24 hours        & \textbf{137.4}                & 688.9                         & 375.5                     \\ \hline
                                                                              & f(w) & 0.1714                        &                                &                               & \textbf{0.1680}               & 0.1682                        & 0.1689                    \\
                                                                              & Acc  & 95.5                          &                                &                               & \textbf{95.6}                 & \textbf{95.6}                 & \textbf{95.6}             \\
                                                                              & nnz  & 96                            &                                &                               & 2654                          & 4499                          & 6050                      \\
\multirow{-4}{*}{\begin{tabular}[c]{@{}l@{}}kdd12\\ C=0.01\end{tabular}}      & Time & 132.1                         & \textgreater{}24 hours         & \textgreater{}24 hours        & 309.4                         & 581.8                         & \textbf{280.4}            \\ \hline
                                                                              & f(w) & 0.1714                        &                                &                               & \textbf{0.1573}               & 0.1612                        & 0.1663                    \\
                                                                              & Acc  & 95.5                          &                                &                               & \textbf{95.6}                 & \textbf{95.6}                 & \textbf{95.6}             \\
                                                                              & nnz  & 103                           &                                &                               & 342325                        & 635187                        & 406605                    \\
\multirow{-4}{*}{\begin{tabular}[c]{@{}l@{}}kdd12\\ C=1\end{tabular}}         & Time & 127.3                         & \textgreater{}24 hours         & \textgreater{}24 hours        & 596.0                         & 639.9                         & \textbf{313.0}            \\ \hline
\end{tabular}
}
\end{table}

A key result is that skglm, Celer, Celer-PN, and SAGA all exhibit three undesirable properties: (1) extraordinary increases in runtime for moderate increases in $C$. (2) Extraordinary increase in runtime for tighter solver tolerances (e.g., skglm went from 34 seconds at $10^{-3}/5$ tolerance to above 24 hours for $10^{-4}$ tolerance). (3) Experience degenerate results in their optimization process (skglm in particular). In most cases, the Celer and SAGA solvers are unable to produce models on large datasets in a reasonable time. The Celer-PN, which uses a ProxyNewton step as an improvement, clearly helps but not to a degree significant enough to change the nature of the results. This speaks to the potential over-design of these methods. 

\textbf{Recommendation:} 
Showing both accuracy and number of non-zeros (nnz) in the solution in \autoref{tbl:main_results} is not included in most prior literature, and exhibits why such reporting should be required to avoid over-designing to unrealistic penalty strengths. As reported in prior work that focuses on a range of $C$, we see that $C =1 $ often outperforms smaller values of $C$, and in most cases $C \geq 1$ achieves the best test-set accuracy\cite{Yuan2012, Yuan2010}. This is also evidenced by $C=0.01$ selecting only a few hundred or thousand out of tens of millions of features. The URL and Webspam datasets in particular, which are the more popular choices in the literature~\cite{bertrand_beyond_2022,massias_dual_2020,pmlr-v80-massias18a}, select less than 100 features at $C=0.01$. This makes sense why the internal workings of skglm start with an initial working-set size of just 10 times. This may play a role in the kdd12, kddb, and Avazu-app datasets where skglm reports far fewer features in use than all other solvers.

The comparatively simpler Liblinear, which uses a 2nd order approximation for the coordinate descent update,
is clearly one of the best performers. Notably, our LBFGS* achieves similar sparsity of solutions and objective function quality $f(\mathbf{w})$ compared to OWL-QN, which is purpose-designed for $L_1$ penalized problems. In many cases, LBFGS* has the best accuracy (8 out of 14), almost always faster than OWL-QN (11/14), and is runtime competitive with Liblinear. The speed improvements compared to OWL-QN are often greatest when the solution is sparse due to our clipping strategy that reduces the amount of work in the LBFGS direction step. 

\begin{figure}[!h]
\begin{center}
\centering
\begin{tikzpicture}[]
\begin{axis}[
    xlabel=Iterations,
    ylabel=Sparsity of Active Set,
    legend style={font=\tiny},
    legend pos=south west,
    legend columns=3,
    xmode=log,
    log basis x={10},
    xmin=1.0,
    height=0.5\columnwidth,
    width=\columnwidth,
    ]

    \addplot[Bittersweet, line width=1.0pt, each nth point={1}, mark=none, solid] table [y expr=1-\thisrow{active_set_size}/3231961, x=iteration, col sep=comma] {data/lbfgs_outs/url_combined_normalized_0.01.csv};
    \addlegendentry{URL}

    \addplot[Bittersweet, line width=1.0pt, each nth point={1}, mark=none, dashed, forget plot] table [y expr=1-\thisrow{active_set_size}/3231961, x=iteration, col sep=comma] {data/lbfgs_outs/url_combined_normalized_1.0.csv};

    \addplot[Blue, line width=1.0pt, each nth point={1}, mark=none, solid] table [y expr=1-\thisrow{active_set_size}/16609143, x=iteration, col sep=comma] {data/lbfgs_outs/webspam_wc_normalized_trigram.svm_0.01.csv};
    \addlegendentry{Webspam}

    \addplot[Blue, line width=1.0pt, each nth point={1}, mark=none, dashed, forget plot] table [y expr=1-\thisrow{active_set_size}/16609143, x=iteration, col sep=comma] {data/lbfgs_outs/webspam_wc_normalized_trigram.svm_1.0.csv};

    \addplot[DarkOrchid, line width=1.0pt, each nth point={1}, mark=none, solid] table [y expr=1-\thisrow{active_set_size}/1000000, x=iteration, col sep=comma] {data/lbfgs_outs/avazu-app.tr_0.01.csv};
    \addlegendentry{Avazu-App}

    \addplot[DarkOrchid, line width=1.0pt, each nth point={1}, mark=none, dashed, forget plot] table [y expr=1-\thisrow{active_set_size}/1000000, x=iteration, col sep=comma] {data/lbfgs_outs/avazu-app.tr_1.0.csv};

    \addplot[Green, line width=1.0pt, each nth point={1}, mark=none, solid] table [y expr=1-\thisrow{active_set_size}/1000000, x=iteration, col sep=comma] {data/lbfgs_outs/avazu-site.tr_0.01.csv};
    \addlegendentry{Avazu-Site}

    \addplot[Green, line width=1.0pt, each nth point={1}, mark=none, dashed, forget plot] table [y expr=1-\thisrow{active_set_size}/1000000, x=iteration, col sep=comma] {data/lbfgs_outs/avazu-site.tr_1.0.csv};

    \addplot[Red, line width=1.0pt, each nth point={1}, mark=none, solid] table [y expr=1-\thisrow{active_set_size}/29890095, x=iteration, col sep=comma] {data/lbfgs_outs/kddb_0.01.csv};
    \addlegendentry{KDDB}

    \addplot[Red, line width=1.0pt, each nth point={1}, mark=none, dashed, forget plot] table [y expr=1-\thisrow{active_set_size}/29890095, x=iteration, col sep=comma] {data/lbfgs_outs/kddb_1.0.csv};

    \addplot[Gray, line width=1.0pt, each nth point={1}, mark=none, solid] table [y expr=1-\thisrow{active_set_size}/54686452, x=iteration, col sep=comma] {data/lbfgs_outs/kdd12.tr_0.01.csv};
    \addlegendentry{KDD12}

    \addplot[Gray, line width=1.0pt, each nth point={1}, mark=none, dashed, forget plot] table [y expr=1-\thisrow{active_set_size}/54686452, x=iteration, col sep=comma] {data/lbfgs_outs/kdd12.tr_1.0.csv};

\end{axis}
\end{tikzpicture}
\end{center}
\caption{Sparsity of the active set (y-axis, higher is better) as optimization iterations proceed (x-axis). Solid lines indicate $C{=}0.01$, dashed lines indicate $C{=}1.0$. In most cases, the majority of dimensions are discarded, enabling faster execution, but in high-density cases the optimization can re-introduce variables periodically.
}
\label{fig:lbfgs_active_set}
\end{figure}
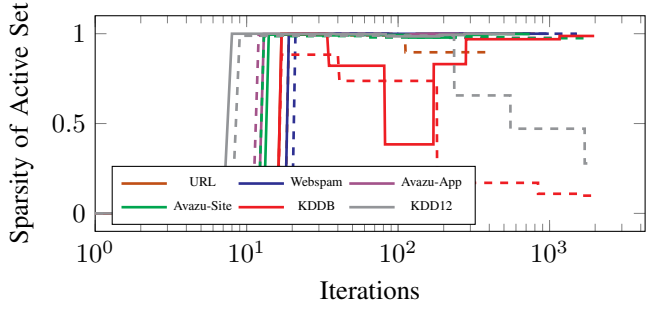

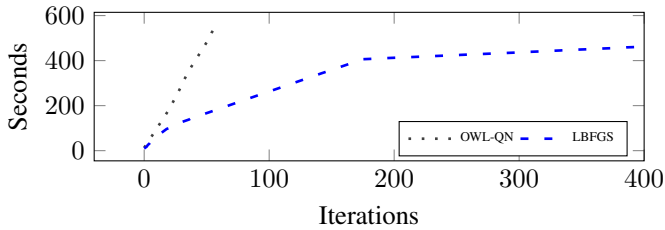
\begin{figure}[!h]
\begin{center}
\centering
\begin{tikzpicture}[]
\begin{axis}[
    xlabel=Iterations,
    ylabel=Seconds,
    legend style={font=\tiny},
    legend pos=south east,
    legend columns=4,
    log basis x={10},
    xmax=400.0,
    height=0.4\columnwidth,
    width=\columnwidth,
    ]

    \addplot[line width=1.0pt,each nth point={2},mark repeat=200,mark=none,loosely dotted,darkgray] table [y=seconds, x=iteration, col sep=comma] {data/owlqn_outs/webspam_wc_normalized_trigram.svm_0.01.csv};
    \addlegendentry{OWL-QN}

    \addplot[line width=1.0pt,each nth point={1},mark repeat=2500,mark=none,loosely dashed,blue] table [y=seconds, x=iteration, col sep=comma] {data/lbfgs_outs/webspam_wc_normalized_trigram.svm_0.01.csv};
    \addlegendentry{LBFGS}
    
\end{axis}
\end{tikzpicture}
\end{center}
\caption{The cumulative amount of time taken (y-axis) as the number of optimization steps increases (x-axis) using the Webspam dataset as an example. Notice the two inflection points for LBFGS*, as the pruning step removes unused dimensions. OWL-QN continues but requires the same level of work at every step, resulting in excess effort. 
}
\label{fig:lbfgs_runtime_clipp}
\end{figure}

The effectiveness of clipping can be seen in \autoref{fig:lbfgs_active_set}, where the sparsity of the active set is plotted against iteration count. The initial active set is all variables and thus starts at 0, and quickly increases after just $\approx 10$ iterations. In the two cases of kddb and kdd12 where the solution is highly dense, we see that the active set also remains large --- but otherwise becomes very small. This leads to significant and increasing speedups over OWL-QN as subsequent iterations are faster than the initial iterations, allowing for more gradient updates in the same time as shown in \autoref{fig:lbfgs_runtime_clipp}. 

Last, we look at the ability to parallelize the solver. Coordinate descent is intrinsically limited in that it iterates one column of features at a time, leaving little to no room for reuse of any values, causing poor memory performance (all accesses of $X$ are cache misses) and inherently sequential calculation. The multi-core version of Liblinear uses the naive approach to parallelizing column calculations from ~\cite{Zhuang:2018:NPC:3269206.3271687} that proports reasonable results for small numbers of processors $p$, but we note the work is now eight years old (with no better since proposed methods that we are aware) and a particularly hardware sensitive approach. In contrast, we apply naive parallelization to the calculation of $X \mathbf{w}$ and $X^\top \nabla f(\mathbf{w})$ that are needed for LBFGS*. The results are shown in \autoref{tbl:speedup}, where we see Liblinear multicore is $\geq 2 \times$ slower when using $p=12$ processors on every test, where LBFGS* has no degenerate cases and usually obtains at least a moderate speedup, making it more reliable from a user perspective of ``no surprises'' in a slowdown. 

\begin{table}[!h]
\caption{Comparison of Liblinear Multi-Core (MC) and LBFGS runtime with naive parallelization for a machine with $p=12$ CPU cores, where \textbf{bold shows multi-code training successfully speeding up the original single-core counterpart}. In all cases, Liblinear is at least $2\times$ slower when 
$p=12$.
} \label{tbl:speedup}
\centering
\adjustbox{max width=0.9\columnwidth}{%
\begin{tabular}{ccrrrr}
\hline
\multicolumn{1}{l}{}         &       & \multicolumn{2}{c}{Liblinear}                           & \multicolumn{2}{c}{LBFGS*}                         \\ \cline{3-4}  \cline{5-6} 
\multicolumn{1}{l}{}         & C     & \multicolumn{1}{c}{p=1} & \multicolumn{1}{c}{p=12}      & \multicolumn{1}{c}{p=1} & \multicolumn{1}{c}{p=12} \\ \hline
                             & 0.001 & 6.7                     & {\color[HTML]{FE0000} 49.8}   & 84.5                    & \textbf{15.1}            \\
\multirow{-2}{*}{URL}        & 1     & 8.7                     & {\color[HTML]{FE0000} 60.6}   & 89.8                    & \textbf{8.5}             \\ \hline
                             & 0.001 & 28.2                    & {\color[HTML]{FE0000} 219.8}  & 448.2                   & \textbf{62.3}            \\
\multirow{-2}{*}{Webspam}    & 1     & 23.6                    & {\color[HTML]{FE0000} 234.0}  & 310.4                   & \textbf{67.1}            \\ \hline
                             & 0.001 & 35.7                    & {\color[HTML]{FE0000} 112.8}  & 30.0                    & \textbf{18.6}            \\
\multirow{-2}{*}{Avazu-app}  & 1     & 20.5                    & {\color[HTML]{FE0000} 63.1}   & 47.1                    & \textbf{6.2}             \\ \hline
                             & 0.001 & 75.1                    & {\color[HTML]{FE0000} 267.3}  & 34.8                    & 35.1                     \\
\multirow{-2}{*}{Avazu-site} & 1     & 42.8                    & {\color[HTML]{FE0000} 175.7}  & 54.9                    & \textbf{39.2}            \\ \hline
                             & 0.001 & 41.5                    & {\color[HTML]{FE0000} 143.8}  & 71.5                    & \textbf{43.2}            \\
\multirow{-2}{*}{KDDA}       & 1     & 65.8                    & {\color[HTML]{FE0000} 215.5}  & 218.1                   & \textbf{180.9}           \\ \hline
                             & 0.001 & 72.0                    & {\color[HTML]{FE0000} 240.5}  & 176.4                   & \textbf{116.4}           \\
\multirow{-2}{*}{KDDB}       & 1     & 137.4                   & {\color[HTML]{FE0000} 448.5}  & 375.5                   & \textbf{372.6}           \\ \hline
                             & 0.001 & 309.4                   & {\color[HTML]{FE0000} 627.7}  & 280.4                   & \textbf{172.1}           \\
\multirow{-2}{*}{KDD12}      & 1     & 596.0                   & {\color[HTML]{FE0000} 1831.2} & 313.0                   & 334.2                    \\ \hline
\end{tabular}
}
\end{table}

This article is not intended to be an exhaustive evaluation of parallel computation, but the result is important in highlighting that the less-explored path of full gradient-based $L_1$ solutions is competitive today and has potential advantages in multi-core and, thus, GPU acceleration. We note that the results are not an aberration and are supported by other observations in the literature. For the speedup of LBFGS*, the calculation of $X^\top \nabla f(\mathbf{w})$ takes half the time and is hard to parallelize in a compressed sparse row format due to intrinsic memory access patterns as noted in ~\cite{lee_fast_2015}. Many alternative file formats are not supported by Scipy/numpy today and could be avenues for future speedup and research~\cite{buluc_parallel_2012,monakov_automatically_2010,liu_efficient_2013, kreutzer_unified_2014,zheng_biell_2014,ballard_hypergraph_2015,filippone_sparse_2017,zhang_vectorized_2018,pogorelov_performance_2023,hwang_cerberus_2024}. These factors also impact distributed $L_1$ solvers that merge the solution to multiple sub-problems, and tradeoffs between how small to break sub-problems up and leverage multi-core solvers on a large machine vs. single-core solvers on more, but smaller, sub-problems~\cite{Izbicki2020,leblond17a,Smith2016,10.1145/3637528.3672038}.

\subsection{Why Does LBFGS* Work?}

It is already known and proven that LBFGS will not converge in a theoretical capacity, and such a result is the genesis for the development of the OWL-QN optimizer~\cite{Gong2015,Andrew2007}. So, why does our method work at all? A proof is, to the best of our knowledge, not possible due to the aforementioned results. However, we can concisely state the engineering insights that led to our method working. 

First, there is the matter of the convergence of the non-zero parameters of the model. Imagine that the set of parameters that are non-zero is the only set of parameters under consideration. The Modified OWL-QN variant we use has three components: 
    (1) A line-search.
    (2) A LBFGS direction calculation (the ``Quasi-Newton'' step).
    (3) A Gradient Descent step that lacks curvature information from the Hessian approximation. 

If the signs of the variables are correct, there is no difference between OWL-QN and LBFGS direction calculations, which we should expect to be true in most instances for non-zero variables. The line search is working to satisfy a sufficiency condition, and so the method of selecting the line search is of no correctness consequence in this instance. That leaves only the third Gradient Descent step, which is required to prove OWL-QN's convergence. Yet, by design, OWL-QN avoids this slower Hessian-free step to ensure the ``QN-step is
adopted at almost all iterations''~\cite{Gong2015}. In this regard, it makes sense that using LBFGS on non-zero variables will yield reasonable convergence, with the risk of failure in cases where ``almost all'' is not satisfied. As shown in our results, this is inconsequential for our practice.

Then, why should we expect our method to work to any reasonable degree for the parameters that should become zero? Prior work in online sparse learning used heuristic approaches to shrink covariates toward zero and simply round sufficiently small weights to zero over multiple iterations~\cite{Langford2009} and using approximations to the $L_1$ penalty for large-scale optimization~\cite{Balakrishnan2008}. Given that these strategies worked successfully, it is not surprising that using LBFGS will push true non-zero gradients \textit{toward} zero. The issue of achieving zeros is circumvented by our pruning step, checking if $|\nabla_j \tilde{\ell}(\mathbf{w}_t)| < \alpha \beta$, which is known to depend on the closeness of the objective function's value, rather than the deviance of the coordinate~\cite{Yuan2010}. Although intertwined, our results, informed by prior literature, demonstrate that LBFGS* is more than sufficient to produce ``close enough'' values, enabling this pruning step to obtain hard zeros without making overly expensive-to-evaluate gradient updates. 

\section{How LBFGS* Impacted Production Use} \label{sec:production_use}

In this section, we detail how LBFGS* addressed multiple production use-case requirements. 

\textbf{Largest Scale, Distributed L1 Optimization Across 30+ TB of data:} As noted above, our target work for this study was to train $L_1$ penalized models over byte n-gram features, where the sample size $\in [3\cdot 10^7, 10^8]$ and dimensionality $d \in [10^7, 10^8]$. This required distributed computation on a Spark cluster to process features and perform training and was primarily I/O-Bound rather than compute-bound. Because each file was stored on a different storage device, each compute node would receive all $d$ features for a subset of the data. Performing coordinate-descent-style training required multiple scatterings of the data across the network, resulting in large performance gaps and slow convergence. Similarly, we found OWL-QN-based solvers insufficient for training at this scale. Our LBGS* approach was the only method we were able to get running reliably and consistently on a 2,0000 CPU core Spark cluster and produce results that we could inspect. 

\textbf{Medium Scale, Single-Powerful Machine:} We had a set of heterogeneous compute machines where each had $\approx$ 128 CPU cores, 4 TB of RAM, and 40 TB of SSD storage. These machines were valuable for dedicating a single powerful compute node to long-running calculations without concern for networking latency and complexity. However, effectively using all CPU cores to train a single $ L_1$-regularized model proved exceptionally challenging. In particular, we performed an extensive parameter search for the penalty parameter $C$, which yielded not only good accuracy but also a balanced selection of non-zero features that met model size and latency targets. We found that one could ``vectorize'' this style of implementation and perform multiple $L_1$ training models with different hyper-parameters $C$ concurrently in a single pass over the data matrix $X$, against a skinny matrix $W= [w_1, w_2, \ldots, w_P]$ for $P$ different processors being used at once. This was faster than using a single-core warm start to iteratively check a sequence $C_{(1)} < C_{(2)} < \ldots < C_{(z)}$, and faster than training multiple LBFGS* or coordinate-descent style solvers concurrently, the naive way. This speedup came because $X$ would often occupy 1-2 TB of RAM in memory, and the process of fetching data itself from memory was the bottleneck; by batch-training in one pass, we minimized data movement and optimized cache and pre-fetching performance. Without this, each model trained naively and independently would be accessing competing portions of $X$ and thrashing the Translation Lookaside Buffer (TLB) cache and the CPU L1,L2, and L3 caches, dramatically reducing throughput by $100\times$. We would often take the best singular parameters found and then train with Liblinear to get the sparsest possible solution at the same accuracy. 

\textbf{Small Scale, $L_1$ Models as A Feature Selector:} The feature selection capabilities of $L_1$ penalties are widely known, but not often used explicitly as a feature selector in the same manner that common scikit-learn feature selectors like forward-backward search or recursive feature elimination are used. However, in our experiments, we found that it was often effective to first train an $ L_1$-penalized model as a feature selector, and then train a subsequent model, such as XGBoost, on the selected features as the final model. This recipe was simple and effective, but it could cause issues with coordinate-descent solvers, which double the memory required to store a column-major-formatted version of the matrix $X$ when most other algorithms expect row-major ordering. With LBFGS*, we build models for downstream users with less difficulty by keeping everything row-major, which was particularly valuable for users who needed to perform regular model training but lacked easy access to high-memory systems.

\section{Conclusion} \label{sec:conclusion}

We have identified a critical issue in the literature of the past decade: using the maximal regularization strength $\lambda_\mathit{max}$ as the basis for a parameter-tuning search, without evaluating the accuracy or sparsity of the solutions found. This has led to multiple papers testing unrealistically sparse solutions and are often incapable of producing timely solutions to reasonable penalty strengths. Not only is an older approach from Liblinear the best all-around coordinate descent-based solver in our testing, but we also find simple modifications to LBFGS are competitive on all large datasets from a practitioner's perspective. These results show the need to improve methodological evaluation and to explore new solvers as a means of improving future work. 

\bibliographystyle{IEEEtran}

\bibliography{references,references-2}

\begin{thebibliography}{10}
\providecommand{\url}[1]{#1}
\csname url@samestyle\endcsname
\providecommand{\newblock}{\relax}
\providecommand{\bibinfo}[2]{#2}
\providecommand{\BIBentrySTDinterwordspacing}{\spaceskip=0pt\relax}
\providecommand{\BIBentryALTinterwordstretchfactor}{4}
\providecommand{\BIBentryALTinterwordspacing}{\spaceskip=\fontdimen2\font plus
\BIBentryALTinterwordstretchfactor\fontdimen3\font minus \fontdimen4\font\relax}
\providecommand{\BIBforeignlanguage}[2]{{%
\expandafter\ifx\csname l@#1\endcsname\relax
\typeout{** WARNING: IEEEtran.bst: No hyphenation pattern has been}%
\typeout{** loaded for the language `#1'. Using the pattern for}%
\typeout{** the default language instead.}%
\else
\language=\csname l@#1\endcsname
\fi
#2}}
\providecommand{\BIBdecl}{\relax}
\BIBdecl

\bibitem{Tibshirani1994}
R.~Tibshirani, ``Regression {Shrinkage} and {Selection} {Via} the {Lasso},'' \emph{Journal of the Royal Statistical Society, Series B}, vol.~58, no.~1, pp. 267--288, 1994.

\bibitem{Fu1998}
W.~J. Fu, ``Penalized {Regressions} : {The} {Bridge} {Versus} the {Lasso},'' \emph{Journal of Computational and Graphical Statistics}, vol.~7, no.~3, pp. 397--416, 1998.

\bibitem{Ng2004}
\BIBentryALTinterwordspacing
A.~Y. Ng, ``Feature selection, {L1} vs. {L2} regularization, and rotational invariance,'' \emph{Twenty-first international conference on Machine learning - ICML '04}, p.~78, 2004, publisher: ACM Press Place: New York, New York, USA ISBN: 1581138285. [Online]. Available: \url{http://portal.acm.org/citation.cfm?doid=1015330.1015435}
\BIBentrySTDinterwordspacing

\bibitem{Zou2005}
\BIBentryALTinterwordspacing
H.~Zou and T.~Hastie, ``Regularization and variable selection via the elastic net,'' \emph{Journal of the Royal Statistical Society, Series B}, vol.~67, no.~2, pp. 301--320, Apr. 2005. [Online]. Available: \url{http://doi.wiley.com/10.1111/j.1467-9868.2005.00503.x}
\BIBentrySTDinterwordspacing

\bibitem{malwareBytes}
\BIBentryALTinterwordspacing
J.~Holt and E.~Raff, ``Malware bytes,'' in \emph{The Next Wave: Cyber Analytics Research}, vol.~25.\hskip 1em plus 0.5em minus 0.4em\relax National Security Agency (NSA), April 2024. [Online]. Available: \url{https://www.govinfo.gov/app/details/GPO-TNW-25-1-2024/GPO-TNW-25-1-2024-6}
\BIBentrySTDinterwordspacing

\bibitem{lu_high-dimensional_2024}
\BIBentryALTinterwordspacing
F.~Lu, R.~R. Curtin, E.~Raff, F.~Ferraro, and J.~Holt, ``High-{Dimensional} {Distributed} {Sparse} {Classification} with {Scalable} {Communication}-{Efficient} {Global} {Updates},'' in \emph{Proceedings of the 30th {ACM} {SIGKDD} {Conference} on {Knowledge} {Discovery} and {Data} {Mining}}, ser. {KDD} '24.\hskip 1em plus 0.5em minus 0.4em\relax New York, NY, USA: Association for Computing Machinery, Aug. 2024, pp. 2037--2047. [Online]. Available: \url{https://dl.acm.org/doi/10.1145/3637528.3672038}
\BIBentrySTDinterwordspacing

\bibitem{raff_saus}
\BIBentryALTinterwordspacing
E.~Raff and J.~Sylvester, ``Linear {Models} with {Many} {Cores} and {CPUs}: {A} {Stochastic} {Atomic} {Update} {Scheme},'' in \emph{2018 {IEEE} {International} {Conference} on {Big} {Data} ({Big} {Data})}.\hskip 1em plus 0.5em minus 0.4em\relax IEEE, Dec. 2018, pp. 65--73. [Online]. Available: \url{https://ieeexplore.ieee.org/document/8622172/}
\BIBentrySTDinterwordspacing

\bibitem{Raff2020autoyara}
\BIBentryALTinterwordspacing
E.~Raff, R.~Zak, G.~L. Munoz, W.~Fleming, H.~S. Anderson, B.~Filar, C.~Nicholas, and J.~Holt, ``Automatic {Yara} {Rule} {Generation} {Using} {Biclustering},'' in \emph{13th {ACM} {Workshop} on {Artificial} {Intelligence} and {Security} ({AISec}’20)}, 2020, arXiv: 2009.03779. [Online]. Available: \url{http://arxiv.org/abs/2009.03779}
\BIBentrySTDinterwordspacing

\bibitem{lu_optimizing_2026}
F.~Lu, R.~R. Curtin, E.~Raff, F.~Ferraro, and J.~Holt, ``\BIBforeignlanguage{en}{Optimizing the {Optimal} {Weighted} {Average}: {Efficient} {Distributed} {Sparse} {Classification}},'' in \emph{\BIBforeignlanguage{en}{Machine {Learning} and {Knowledge} {Discovery} in {Databases}. {Research} {Track}}}, R.~P. Ribeiro, B.~Pfahringer, N.~Japkowicz, P.~Larrañaga, A.~M. Jorge, C.~Soares, P.~H. Abreu, and J.~Gama, Eds.\hskip 1em plus 0.5em minus 0.4em\relax Cham: Springer Nature Switzerland, 2026, pp. 147--163.

\bibitem{raff_ngram_2016}
\BIBentryALTinterwordspacing
E.~Raff, R.~Zak, R.~Cox, J.~Sylvester, P.~Yacci, R.~Ward, A.~Tracy, M.~McLean, and C.~Nicholas, ``An investigation of byte n-gram features for malware classification,'' \emph{Journal of Computer Virology and Hacking Techniques}, Sep. 2016. [Online]. Available: \url{http://link.springer.com/10.1007/s11416-016-0283-1}
\BIBentrySTDinterwordspacing

\bibitem{Kilograms_2019}
\BIBentryALTinterwordspacing
E.~Raff, W.~Fleming, R.~Zak, H.~Anderson, B.~Finlayson, C.~K. Nicholas, M.~Mclean, W.~Fleming, C.~K. Nicholas, R.~Zak, and M.~Mclean, ``{KiloGrams}: {Very} {Large} {N}-{Grams} for {Malware} {Classification},'' in \emph{Proceedings of {KDD} 2019 {Workshop} on {Learning} and {Mining} for {Cybersecurity} ({LEMINCS}’19)}, 2019. [Online]. Available: \url{https://arxiv.org/abs/1908.00200}
\BIBentrySTDinterwordspacing

\bibitem{raff_zipf-gramming_2026}
\BIBentryALTinterwordspacing
E.~Raff, R.~R. Curtin, D.~Everett, R.~J. Joyce, and J.~Holt, ``Zipf-{Gramming}: {Scaling} {Byte} {N}-{Grams} {Up} to {Production} {Sized} {Malware} {Corpora},'' in \emph{Proceedings of the 34th {ACM} {International} {Conference} on {Information} and {Knowledge} {Management}}, ser. {CIKM} '25.\hskip 1em plus 0.5em minus 0.4em\relax New York, NY, USA: Association for Computing Machinery, 2026, pp. 5988--5996. [Online]. Available: \url{https://doi.org/10.1145/3746252.3761551}
\BIBentrySTDinterwordspacing

\bibitem{curtin_intermediate_2025}
\BIBentryALTinterwordspacing
R.~R. Curtin, F.~Lu, E.~Raff, and P.~Ranade, ``Intermediate {N}-{Gramming}: {Deterministic} and {Fast} {N}-{Grams} {For} {Large} {N} and {Large} {Datasets},'' in \emph{Proceedings of the {AAAI} {Conference} on {Artificial} {Intelligence}}, vol.~40.\hskip 1em plus 0.5em minus 0.4em\relax arXiv, Nov. 2025, arXiv:2511.14955 [cs]. [Online]. Available: \url{http://arxiv.org/abs/2511.14955}
\BIBentrySTDinterwordspacing

\bibitem{Harang2020}
\BIBentryALTinterwordspacing
R.~Harang and E.~M. Rudd, ``{SOREL}-{20M}: {A} {Large} {Scale} {Benchmark} {Dataset} for {Malicious} {PE} {Detection},'' \emph{arXiv}, 2020, arXiv: 2012.07634. [Online]. Available: \url{http://arxiv.org/abs/2012.07634}
\BIBentrySTDinterwordspacing

\bibitem{Joyce2022}
\BIBentryALTinterwordspacing
R.~J. Joyce, D.~Amlani, C.~Nicholas, and E.~Raff, ``{MOTIF}: {A} {Large} {Malware} {Reference} {Dataset} with {Ground} {Truth} {Family} {Labels},'' in \emph{The {AAAI}-22 {Workshop} on {Artificial} {Intelligence} for {Cyber} {Security} ({AICS})}, 2022, arXiv: 2111.15031v1. [Online]. Available: \url{https://github.com/boozallen/MOTIF}
\BIBentrySTDinterwordspacing

\bibitem{joyce_ember2024_2025}
\BIBentryALTinterwordspacing
R.~J. Joyce, G.~Miller, P.~Roth, R.~Zak, E.~Zaresky-Williams, H.~Anderson, E.~Raff, and J.~Holt, ``{EMBER2024} - {A} {Benchmark} {Dataset} for {Holistic} {Evaluation} of {Malware} {Classifiers},'' in \emph{Proceedings of the 31st {ACM} {SIGKDD} {Conference} on {Knowledge} {Discovery} and {Data} {Mining} {V}.2}, ser. {KDD} '25.\hskip 1em plus 0.5em minus 0.4em\relax New York, NY, USA: Association for Computing Machinery, Aug. 2025, pp. 5516--5526. [Online]. Available: \url{https://dl.acm.org/doi/10.1145/3711896.3737431}
\BIBentrySTDinterwordspacing

\bibitem{raff_what_2025}
\BIBentryALTinterwordspacing
E.~Raff, M.~Benaroch, S.~Samtani, and A.~L. Farris, ``\BIBforeignlanguage{en}{What {Do} {Machine} {Learning} {Researchers} {Mean} by “{Reproducible}”?}'' \emph{\BIBforeignlanguage{en}{Proceedings of the AAAI Conference on Artificial Intelligence}}, vol.~39, no.~27, pp. 28\,671--28\,683, Apr. 2025. [Online]. Available: \url{https://ojs.aaai.org/index.php/AAAI/article/view/35093}
\BIBentrySTDinterwordspacing

\bibitem{Raff2022a}
\BIBentryALTinterwordspacing
E.~Raff and A.~L. Farris, ``A {Siren} {Song} of {Open} {Source} {Reproducibility},'' in \emph{{ML} {Evaluation} {Standards} {Workshop} at {ICLR} 2022}, 2022. [Online]. Available: \url{https://arxiv.org/abs/2204.04372}
\BIBentrySTDinterwordspacing

\bibitem{raff_reproducibility_2023}
\BIBentryALTinterwordspacing
E.~Raff and J.~Holt, ``\BIBforeignlanguage{en}{Reproducibility in {Multiple} {Instance} {Learning}: {A} {Case} {For} {Algorithmic} {Unit} {Tests}},'' \emph{\BIBforeignlanguage{en}{Advances in Neural Information Processing Systems}}, vol.~36, pp. 13\,530--13\,544, Dec. 2023. [Online]. Available: \url{https://proceedings.neurips.cc/paper_files/paper/2023/hash/2bab8865fa4511e445767e3750b2b5ac-Abstract-Conference.html}
\BIBentrySTDinterwordspacing

\bibitem{Raff2022}
\BIBentryALTinterwordspacing
E.~Raff, ``Does the {Market} of {Citations} {Reward} {Reproducible} {Work}?'' in \emph{{ML} {Evaluation} {Standards} {Workshop} at {ICLR} 2022}, 2022. [Online]. Available: \url{https://arxiv.org/abs/2204.03829}
\BIBentrySTDinterwordspacing

\bibitem{Raff2020c}
\BIBentryALTinterwordspacing
------, ``Research {Reproducibility} as a {Survival} {Analysis},'' in \emph{The {Thirty}-{Fifth} {AAAI} {Conference} on {Artificial} {Intelligence}}, 2021, arXiv: 2012.09932. [Online]. Available: \url{http://arxiv.org/abs/2012.09932}
\BIBentrySTDinterwordspacing

\bibitem{Raff2019_quantify_repro}
\BIBentryALTinterwordspacing
------, ``A {Step} {Toward} {Quantifying} {Independently} {Reproducible} {Machine} {Learning} {Research},'' in \emph{{NeurIPS}}, 2019, arXiv: 1909.06674. [Online]. Available: \url{http://arxiv.org/abs/1909.06674}
\BIBentrySTDinterwordspacing

\bibitem{Musgrave2020}
\BIBentryALTinterwordspacing
K.~Musgrave, S.~Belongie, and S.-N. Lim, ``A {Metric} {Learning} {Reality} {Check},'' in \emph{{ECCV}}, 2020, arXiv: 2003.08505. [Online]. Available: \url{http://arxiv.org/abs/2003.08505}
\BIBentrySTDinterwordspacing

\bibitem{coakley_examining_2022}
K.~Coakley, C.~R. Kirkpatrick, and O.~E. Gundersen, ``Examining the {Effect} of {Implementation} {Factors} on {Deep} {Learning} {Reproducibility},'' in \emph{2022 {IEEE} 18th {International} {Conference} on e-{Science} (e-{Science})}, Oct. 2022, pp. 397--398.

\bibitem{Gundersen2018}
O.~E. Gundersen and S.~Kjensmo, ``State of the {Art}: {Reproducibility} in {Artificial} {Intelligence},'' \emph{Proceedings of the 32nd AAAI Conference on Artificial Intelligence (AAAI-18)}, pp. 1644--1651, 2018.

\bibitem{Friedman2010}
J.~Friedman, T.~Hastie, and R.~Tibshirani, ``Regularization {Paths} for {Generalized} {Linear} {Models} via {Coordinate} {Descent},'' \emph{Journal of Statistical Software}, vol.~33, no.~1, pp. 1--22, 2010, arXiv: 1501.0228 ISBN: 9781439811870.

\bibitem{Yuan2012}
\BIBentryALTinterwordspacing
G.-x. Yuan, C.-H. Ho, and C.-j. Lin, ``An improved {GLMNET} for {L1}-regularized logistic regression,'' \emph{Journal of Machine Learning Research}, vol.~13, pp. 1999--2030, 2012, publisher: ACM Press Place: New York, New York, USA ISBN: 9781450308137. [Online]. Available: \url{http://dl.acm.org/citation.cfm?doid=2020408.2020421}
\BIBentrySTDinterwordspacing

\bibitem{Fan2008}
R.-E. Fan, K.-W. Chang, C.-J. Hsieh, X.-R. Wang, and C.-J. Lin, ``{LIBLINEAR}: {A} {Library} for {Large} {Linear} {Classification},'' \emph{The Journal of Machine Learning Research}, vol.~9, pp. 1871--1874, 2008.

\bibitem{ElGhaoui2010}
L.~El~Ghaoui, V.~Viallon, and T.~Rabbani, ``Safe feature elimination in sparse supervised learning,'' \emph{Pacific Journal of Optimization}, vol.~8, pp. 667--698, 2010, arXiv: 1009.4219.

\bibitem{Tibshirani2012}
R.~Tibshirani, J.~Bien, J.~Friedman, T.~Hastie, N.~Simon, J.~Taylor, and R.~J. Tibshirani, ``{Strong rules for discarding predictors in lasso-type problems},'' \emph{Journal of the Royal Statistical Society. Series B: Statistical Methodology}, vol.~74, no.~1, pp. 245--266, 2012.

\bibitem{wang_lasso_2013}
\BIBentryALTinterwordspacing
J.~Wang, J.~Zhou, P.~Wonka, and J.~Ye, ``Lasso {Screening} {Rules} via {Dual} {Polytope} {Projection},'' in \emph{Advances in {Neural} {Information} {Processing} {Systems}}, vol.~26.\hskip 1em plus 0.5em minus 0.4em\relax Curran Associates, Inc., 2013. [Online]. Available: \url{https://papers.nips.cc/paper/2013/hash/8b16ebc056e613024c057be590b542eb-Abstract.html}
\BIBentrySTDinterwordspacing

\bibitem{Wang2014a}
J.~Wang, J.~Zhou, J.~Liu, P.~Wonka, and J.~Ye, ``A {Safe} {Screening} {Rule} for {Sparse} {Logistic} {Regression},'' in \emph{Advances in {Neural} {Information} {Processing} {Systems} 27}, Z.~Ghahramani, M.~Welling, C.~Cortes, N.~Lawrence, and K.~Weinberger, Eds.\hskip 1em plus 0.5em minus 0.4em\relax Curran Associates, Inc., 2014, pp. 1053--1061, arXiv: 1307.4145v2.

\bibitem{ndiaye_gap_2017}
\BIBentryALTinterwordspacing
E.~Ndiaye, O.~Fercoq, Alex, R.~Gramfort, and J.~Salmon, ``Gap {Safe} {Screening} {Rules} for {Sparsity} {Enforcing} {Penalties},'' \emph{Journal of Machine Learning Research}, vol.~18, no. 128, pp. 1--33, 2017. [Online]. Available: \url{http://jmlr.org/papers/v18/16-577.html}
\BIBentrySTDinterwordspacing

\bibitem{rakotomamonjy_screening_2019}
\BIBentryALTinterwordspacing
A.~Rakotomamonjy, G.~Gasso, and J.~Salmon, ``\BIBforeignlanguage{en}{Screening rules for {Lasso} with non-convex {Sparse} {Regularizers}},'' in \emph{\BIBforeignlanguage{en}{Proceedings of the 36th {International} {Conference} on {Machine} {Learning}}}.\hskip 1em plus 0.5em minus 0.4em\relax PMLR, May 2019, pp. 5341--5350, iSSN: 2640-3498. [Online]. Available: \url{https://proceedings.mlr.press/v97/rakotomamonjy19a.html}
\BIBentrySTDinterwordspacing

\bibitem{dantas_expanding_2021}
\BIBentryALTinterwordspacing
C.~F. Dantas, E.~Soubies, and C.~Févotte, ``Expanding {Boundaries} of {Gap} {Safe} {Screening},'' \emph{Journal of Machine Learning Research}, vol.~22, no. 236, pp. 1--57, 2021. [Online]. Available: \url{http://jmlr.org/papers/v22/21-0179.html}
\BIBentrySTDinterwordspacing

\bibitem{Larsson2021}
\BIBentryALTinterwordspacing
J.~Larsson, ``Look-{Ahead} {Screening} {Rules} for the {Lasso},'' 2021, arXiv: 2105.05648. [Online]. Available: \url{http://arxiv.org/abs/2105.05648}
\BIBentrySTDinterwordspacing

\bibitem{Lin2008}
C.-j. Lin, R.~C. Weng, and S.~S. Keerthi, ``Trust {Region} {Newton} {Method} for {Large}-{Scale} {Logistic} {Regression},'' \emph{The Journal of Machine Learning Research}, vol.~9, pp. 627--650, 2008.

\bibitem{bertrand_anderson_2021}
\BIBentryALTinterwordspacing
Q.~Bertrand and M.~Massias, ``\BIBforeignlanguage{en}{Anderson acceleration of coordinate descent},'' in \emph{\BIBforeignlanguage{en}{Proceedings of {The} 24th {International} {Conference} on {Artificial} {Intelligence} and {Statistics}}}.\hskip 1em plus 0.5em minus 0.4em\relax PMLR, Mar. 2021, pp. 1288--1296, iSSN: 2640-3498. [Online]. Available: \url{https://proceedings.mlr.press/v130/bertrand21a.html}
\BIBentrySTDinterwordspacing

\bibitem{bertrand_beyond_2022}
\BIBentryALTinterwordspacing
Q.~Bertrand, Q.~Klopfenstein, P.-A. Bannier, G.~Gidel, and M.~Massias, ``\BIBforeignlanguage{en}{Beyond {L1}: {Faster} and {Better} {Sparse} {Models} with skglm},'' \emph{\BIBforeignlanguage{en}{Advances in Neural Information Processing Systems}}, vol.~35, pp. 38\,950--38\,965, Dec. 2022. [Online]. Available: \url{https://proceedings.neurips.cc/paper_files/paper/2022/hash/fe5c31e525e9a26a1426ab0b589f42fe-Abstract-Conference.html}
\BIBentrySTDinterwordspacing

\bibitem{pmlr-v37-johnson15}
\BIBentryALTinterwordspacing
T.~Johnson and C.~Guestrin, ``Blitz: {A} {Principled} {Meta}-{Algorithm} for {Scaling} {Sparse} {Optimization},'' in \emph{Proceedings of the 32nd {International} {Conference} on {Machine} {Learning}}, F.~Bach and D.~Blei, Eds., vol.~37.\hskip 1em plus 0.5em minus 0.4em\relax Lille, France: PMLR, 2015, pp. 1171--1179, series Title: Proceedings of Machine Learning Research. [Online]. Available: \url{http://proceedings.mlr.press/v37/johnson15.html}
\BIBentrySTDinterwordspacing

\bibitem{rakotomamonjy_convergent_2022}
\BIBentryALTinterwordspacing
A.~Rakotomamonjy, R.~Flamary, J.~Salmon, and G.~Gasso, ``\BIBforeignlanguage{en}{Convergent {Working} {Set} {Algorithm} for {Lasso} with {Non}-{Convex} {Sparse} {Regularizers}},'' in \emph{\BIBforeignlanguage{en}{Proceedings of {The} 25th {International} {Conference} on {Artificial} {Intelligence} and {Statistics}}}.\hskip 1em plus 0.5em minus 0.4em\relax PMLR, May 2022, pp. 5196--5211, iSSN: 2640-3498. [Online]. Available: \url{https://proceedings.mlr.press/v151/rakotomamonjy22a.html}
\BIBentrySTDinterwordspacing

\bibitem{NIPS2014_5258}
A.~Defazio, F.~Bach, and S.~Lacoste-Julien, ``{SAGA}: {A} {Fast} {Incremental} {Gradient} {Method} {With} {Support} for {Non}-{Strongly} {Convex} {Composite} {Objectives},'' in \emph{Advances in {Neural} {Information} {Processing} {Systems} 27}, Z.~Ghahramani, M.~Welling, C.~Cortes, N.~D. Lawrence, and K.~Q. Weinberger, Eds.\hskip 1em plus 0.5em minus 0.4em\relax Curran Associates, Inc., 2014, pp. 1646--1654.

\bibitem{scikit-learn}
\BIBentryALTinterwordspacing
F.~Pedregosa, G.~Varoquaux, A.~Gramfort, V.~Michel, B.~Thirion, O.~Grisel, M.~Blondel, P.~Prettenhofer, R.~Weiss, V.~Dubourg, J.~Vanderplas, A.~Passos, D.~Cournapeau, M.~Brucher, M.~Perrot, and E.~Duchesnay, ``Scikit-learn: {Machine} {Learning} in {Python},'' \emph{Journal of Machine Learning Research}, vol.~12, pp. 2825--2830, 2011. [Online]. Available: \url{http://jmlr.csail.mit.edu/papers/v12/pedregosa11a.html}
\BIBentrySTDinterwordspacing

\bibitem{Andrew2007}
\BIBentryALTinterwordspacing
G.~Andrew and J.~Gao, ``Scalable training of {L1} -regularized log-linear models,'' in \emph{Proceedings of the 24th international conference on {Machine} learning - {ICML} '07}.\hskip 1em plus 0.5em minus 0.4em\relax New York, New York, USA: ACM Press, 2007, pp. 33--40. [Online]. Available: \url{http://portal.acm.org/citation.cfm?doid=1273496.1273501}
\BIBentrySTDinterwordspacing

\bibitem{Gong2015}
P.~Gong and J.~Ye, ``A {Modified} {Orthant}-{Wise} {Limited} {Memory} {Quasi}-{Newton} {Method} with {Convergence} {Analysis},'' in \emph{The 32nd {International} {Conference} on {Machine} {Learning}}, vol.~37, 2015.

\bibitem{10882922}
A.~Bemporad, ``An l-bfgs-b approach for linear and nonlinear system identification under $\ell _{1}$ and group-lasso regularization,'' \emph{IEEE Transactions on Automatic Control}, pp. 1--8, 2025.

\bibitem{Zhou2015}
Q.~Zhou, W.~Chen, S.~Song, J.~R. Gardner, K.~Q. Weinberger, and Y.~Chen, ``A {Reduction} of the {Elastic} {Net} to {Support} {Vector} {Machines} with an {Application} to {GPU} {Computing},'' in \emph{Association for the {Advancement} of {Artificial} {Intelligence}}, 2015.

\bibitem{ziyin_spred_2023}
\BIBentryALTinterwordspacing
L.~Ziyin and Z.~Wang, ``\BIBforeignlanguage{en}{spred: {Solving} {L1} {Penalty} with {SGD}},'' in \emph{\BIBforeignlanguage{en}{Proceedings of the 40th {International} {Conference} on {Machine} {Learning}}}.\hskip 1em plus 0.5em minus 0.4em\relax PMLR, Jul. 2023, pp. 43\,407--43\,422, iSSN: 2640-3498. [Online]. Available: \url{https://proceedings.mlr.press/v202/ziyin23a.html}
\BIBentrySTDinterwordspacing

\bibitem{goldstein_split_2009}
\BIBentryALTinterwordspacing
T.~Goldstein and S.~Osher, ``The {Split} {Bregman} {Method} for {L1}-{Regularized} {Problems},'' \emph{SIAM Journal on Imaging Sciences}, vol.~2, no.~2, pp. 323--343, Jan. 2009, publisher: Society for Industrial and Applied Mathematics. [Online]. Available: \url{https://epubs.siam.org/doi/10.1137/080725891}
\BIBentrySTDinterwordspacing

\bibitem{koh_method_2007}
K.~Koh, S.-J. Kim, and S.~Boyd, ``A method for large-scale l1-regularized logistic regression,'' in \emph{Proceedings of the 22nd national conference on {Artificial} intelligence - {Volume} 1}, ser. {AAAI}'07.\hskip 1em plus 0.5em minus 0.4em\relax Vancouver, British Columbia, Canada: AAAI Press, Jul. 2007, pp. 565--571.

\bibitem{shi_fast_2010}
\BIBentryALTinterwordspacing
J.~Shi, W.~Yin, S.~Osher, and P.~Sajda, ``A {Fast} {Hybrid} {Algorithm} for {Large}-{Scale} \textit{l$_{\textrm{1}}$}-{Regularized} {Logistic} {Regression},'' \emph{Journal of Machine Learning Research}, vol.~11, no.~23, pp. 713--741, 2010. [Online]. Available: \url{http://jmlr.org/papers/v11/shi10a.html}
\BIBentrySTDinterwordspacing

\bibitem{dai_rehline_2023}
\BIBentryALTinterwordspacing
B.~Dai and Y.~Qiu, ``\BIBforeignlanguage{en}{{ReHLine}: {Regularized} {Composite} {ReLU}-{ReHU} {Loss} {Minimization} with {Linear} {Computation} and {Linear} {Convergence}},'' \emph{\BIBforeignlanguage{en}{Advances in Neural Information Processing Systems}}, vol.~36, pp. 14\,366--14\,386, Dec. 2023. [Online]. Available: \url{https://proceedings.neurips.cc/paper_files/paper/2023/hash/2e37e56599e3f49cc899f40ae4f5d1fa-Abstract-Conference.html}
\BIBentrySTDinterwordspacing

\bibitem{blondel_lightning_2016}
\BIBentryALTinterwordspacing
M.~Blondel and F.~Pedregosa, ``Lightning: large-scale linear classification, regression and ranking in {Python},'' Dec. 2016. [Online]. Available: \url{https://zenodo.org/records/200504}
\BIBentrySTDinterwordspacing

\bibitem{massias_dual_2020}
\BIBentryALTinterwordspacing
M.~Massias, S.~Vaiter, A.~Gramfort, and J.~Salmon, ``Dual {Extrapolation} for {Sparse} {GLMs},'' \emph{Journal of Machine Learning Research}, vol.~21, no. 234, pp. 1--33, 2020. [Online]. Available: \url{http://jmlr.org/papers/v21/19-587.html}
\BIBentrySTDinterwordspacing

\bibitem{pmlr-v80-massias18a}
\BIBentryALTinterwordspacing
M.~Massias, J.~Salmon, and A.~Gramfort, ``Celer: a {Fast} {Solver} for the {Lasso} with {Dual} {Extrapolation},'' in \emph{Proceedings of the 35th {International} {Conference} on {Machine} {Learning}}, J.~Dy and A.~Krause, Eds., vol.~80.\hskip 1em plus 0.5em minus 0.4em\relax Stockholmsmässan, Stockholm Sweden: PMLR, 2018, pp. 3321--3330, series Title: Proceedings of Machine Learning Research. [Online]. Available: \url{http://proceedings.mlr.press/v80/massias18a.html}
\BIBentrySTDinterwordspacing

\bibitem{barroso-luque_sparse-lm_2023}
\BIBentryALTinterwordspacing
L.~Barroso-Luque and F.~Xie, ``\BIBforeignlanguage{en}{sparse-lm: {Sparse} linear regression models in {Python}},'' \emph{\BIBforeignlanguage{en}{Journal of Open Source Software}}, vol.~8, no.~92, p. 5867, Dec. 2023. [Online]. Available: \url{https://joss.theoj.org/papers/10.21105/joss.05867}
\BIBentrySTDinterwordspacing

\bibitem{ren_thunder_2020}
\BIBentryALTinterwordspacing
S.~Ren, W.~Zhao, and P.~Li, ``Thunder: a {Fast} {Coordinate} {Selection} {Solver} for {Sparse} {Learning},'' in \emph{Advances in {Neural} {Information} {Processing} {Systems}}, vol.~33.\hskip 1em plus 0.5em minus 0.4em\relax Curran Associates, Inc., 2020, pp. 1571--1582. [Online]. Available: \url{https://proceedings.neurips.cc/paper/2020/hash/11348e03e23b137d55d94464250a67a2-Abstract.html}
\BIBentrySTDinterwordspacing

\bibitem{zeng_biglasso_2018}
\BIBentryALTinterwordspacing
Y.~Zeng and P.~Breheny, ``The biglasso {Package}: {A} {Memory}- and {Computation}-{Efficient} {Solver} for {Lasso} {Model} {Fitting} with {Big} {Data} in {R},'' Mar. 2018, arXiv:1701.05936 [stat]. [Online]. Available: \url{http://arxiv.org/abs/1701.05936}
\BIBentrySTDinterwordspacing

\bibitem{kim_another_2018}
\BIBentryALTinterwordspacing
D.~Kim and J.~A. Fessler, ``Another {Look} at the {Fast} {Iterative} {Shrinkage}/{Thresholding} {Algorithm} ({FISTA}),'' \emph{SIAM Journal on Optimization}, vol.~28, no.~1, pp. 223--250, Jan. 2018, publisher: Society for Industrial and Applied Mathematics. [Online]. Available: \url{https://epubs.siam.org/doi/10.1137/16M108940X}
\BIBentrySTDinterwordspacing

\bibitem{moreau_benchopt_2022}
\BIBentryALTinterwordspacing
T.~Moreau, M.~Massias, A.~Gramfort, P.~Ablin, P.-A. Bannier, B.~Charlier, M.~Dagréou, T.~D. la~Tour, G.~Durif, C.~F. Dantas, Q.~Klopfenstein, J.~Larsson, E.~Lai, T.~Lefort, B.~Malézieux, B.~Moufad, B.~T. Nguyen, A.~Rakotomamonjy, Z.~Ramzi, J.~Salmon, and S.~Vaiter, ``Benchopt: {Reproducible}, efficient and collaborative optimization benchmarks,'' Oct. 2022, arXiv:2206.13424 [cs, math, stat]. [Online]. Available: \url{http://arxiv.org/abs/2206.13424}
\BIBentrySTDinterwordspacing

\bibitem{Yang:2019:ENE:3292500.3330910}
\BIBentryALTinterwordspacing
S.~Yang, J.~Wen, X.~Zhan, and D.~Kifer, ``{ET}-{Lasso}: {A} {New} {Efficient} {Tuning} of {Lasso}-type {Regularization} for {High}-{Dimensional} {Data},'' in \emph{Proceedings of the 25th {ACM} {SIGKDD} {International} {Conference} on {Knowledge} {Discovery} \& {Data} {Mining}}.\hskip 1em plus 0.5em minus 0.4em\relax New York, NY, USA: ACM, 2019, pp. 607--616, series Title: KDD '19. [Online]. Available: \url{http://doi.acm.org/10.1145/3292500.3330910}
\BIBentrySTDinterwordspacing

\bibitem{liu_large-scale_2009}
\BIBentryALTinterwordspacing
J.~Liu, J.~Chen, and J.~Ye, ``Large-scale sparse logistic regression,'' in \emph{Proceedings of the 15th {ACM} {SIGKDD} international conference on {Knowledge} discovery and data mining}, ser. {KDD} '09.\hskip 1em plus 0.5em minus 0.4em\relax New York, NY, USA: Association for Computing Machinery, Jun. 2009, pp. 547--556. [Online]. Available: \url{https://dl.acm.org/doi/10.1145/1557019.1557082}
\BIBentrySTDinterwordspacing

\bibitem{4407762}
M.~A.~T. Figueiredo, R.~D. Nowak, and S.~J. Wright, ``Gradient projection for sparse reconstruction: Application to compressed sensing and other inverse problems,'' \emph{IEEE Journal of Selected Topics in Signal Processing}, vol.~1, no.~4, pp. 586--597, 2007.

\bibitem{Figueiredo2003}
\BIBentryALTinterwordspacing
M.~A. Figueiredo, ``{ Adaptive Sparseness for Supervised Learning },'' \emph{IEEE Transactions on Pattern Analysis \& Machine Intelligence}, vol.~25, no.~09, pp. 1150--1159, Sep. 2003. [Online]. Available: \url{https://doi.ieeecomputersociety.org/10.1109/TPAMI.2003.1227989}
\BIBentrySTDinterwordspacing

\bibitem{Lawrence_etal_2004}
\BIBentryALTinterwordspacing
L.~Carin, M.~A. Figueiredo, B.~Krishnapuram, and A.~J. Hartemink, ``{ A Bayesian Approach to Joint Feature Selection and Classifier Design },'' \emph{IEEE Transactions on Pattern Analysis \& Machine Intelligence}, vol.~26, no.~09, pp. 1105--1111, Sep. 2004. [Online]. Available: \url{https://doi.ieeecomputersociety.org/10.1109/TPAMI.2004.55}
\BIBentrySTDinterwordspacing

\bibitem{Lawrence_etal_2005}
\BIBentryALTinterwordspacing
L.~Carin, A.~J. Hartemink, B.~Krishnapuram, and M.~A. Figueiredo, ``{ Sparse Multinomial Logistic Regression: Fast Algorithms and Generalization Bounds },'' \emph{IEEE Transactions on Pattern Analysis \& Machine Intelligence}, vol.~27, no.~06, pp. 957--968, Jun. 2005. [Online]. Available: \url{https://doi.ieeecomputersociety.org/10.1109/TPAMI.2005.127}
\BIBentrySTDinterwordspacing

\bibitem{Yuan2010}
G.-X. Yuan, K.-W. Chang, C.-J. Hsieh, and C.-J. Lin, ``{A Comparison of Optimization Methods and Software for Large-scale L1-regularized Linear Classification},'' \emph{The Journal of Machine Learning Research}, vol.~11, pp. 3183--3234, 2010.

\bibitem{Brent1971}
\BIBentryALTinterwordspacing
R.~P. Brent, ``An algorithm with guaranteed convergence for finding a zero of a function,'' \emph{The Computer Journal}, vol.~14, no.~4, pp. 422--425, 1971, iSBN: 0-13-022335-2. [Online]. Available: \url{https://academic.oup.com/comjnl/article-lookup/doi/10.1093/comjnl/14.4.422}
\BIBentrySTDinterwordspacing

\bibitem{Armijo1966}
\BIBentryALTinterwordspacing
L.~Armijo, ``Minimization of functions having lipschitz continuous first partial derivatives,'' \emph{Pacific Journal of Mathematics}, vol.~16, no.~1, p. 1–3, Jan. 1966. [Online]. Available: \url{http://dx.doi.org/10.2140/pjm.1966.16.1}
\BIBentrySTDinterwordspacing

\bibitem{Chang2011}
\BIBentryALTinterwordspacing
C.-C. Chang and C.-J. Lin, ``{LIBSVM}: {A} library for support vector machines,'' \emph{ACM Transactions on Intelligent Systems and Technology}, vol.~2, no.~3, Apr. 2011. [Online]. Available: \url{http://dl.acm.org/citation.cfm?doid=1961189.1961199}
\BIBentrySTDinterwordspacing

\bibitem{Platt1998}
J.~C. Platt, ``Sequential {Minimal} {Optimization}: {A} {Fast} {Algorithm} for {Training} {Support} {Vector} {Machines},'' in \emph{Advances in kernel methods}, 1998, pp. 185--208.

\bibitem{Abeel2009}
T.~Abeel, Y.~V.~D. Peer, and Y.~Saeys, ``Java-{ML}: {A} {Machine} {Learning} {Library},'' \emph{Journal of Machine Learning Research}, vol.~10, pp. 931--934, 2009, iSBN: 1532-4435.

\bibitem{Zhuang:2018:NPC:3269206.3271687}
\BIBentryALTinterwordspacing
Y.~Zhuang, Y.~Juan, G.-X. Yuan, and C.-J. Lin, ``Naive {Parallelization} of {Coordinate} {Descent} {Methods} and an {Application} on {Multi}-core {L1}-regularized {Classification},'' in \emph{Proceedings of the 27th {ACM} {International} {Conference} on {Information} and {Knowledge} {Management}}.\hskip 1em plus 0.5em minus 0.4em\relax New York, NY, USA: ACM, 2018, pp. 1103--1112, series Title: CIKM '18. [Online]. Available: \url{http://doi.acm.org/10.1145/3269206.3271687}
\BIBentrySTDinterwordspacing

\bibitem{lee_fast_2015}
\BIBentryALTinterwordspacing
M.-C. Lee, W.-L. Chiang, and C.-J. Lin, ``Fast {Matrix}-{Vector} {Multiplications} for {Large}-{Scale} {Logistic} {Regression} on {Shared}-{Memory} {Systems},'' in \emph{2015 {IEEE} {International} {Conference} on {Data} {Mining}}, Nov. 2015, pp. 835--840, iSSN: 1550-4786. [Online]. Available: \url{https://ieeexplore.ieee.org/document/7373398}
\BIBentrySTDinterwordspacing

\bibitem{buluc_parallel_2012}
\BIBentryALTinterwordspacing
A.~Buluç and J.~R. Gilbert, ``Parallel {Sparse} {Matrix}-{Matrix} {Multiplication} and {Indexing}: {Implementation} and {Experiments},'' \emph{SIAM Journal on Scientific Computing}, vol.~34, no.~4, pp. C170--C191, Jan. 2012, publisher: Society for Industrial and Applied Mathematics. [Online]. Available: \url{https://epubs.siam.org/doi/10.1137/110848244}
\BIBentrySTDinterwordspacing

\bibitem{monakov_automatically_2010}
A.~Monakov, A.~Lokhmotov, and A.~Avetisyan, ``\BIBforeignlanguage{en}{Automatically {Tuning} {Sparse} {Matrix}-{Vector} {Multiplication} for {GPU} {Architectures}},'' in \emph{\BIBforeignlanguage{en}{High {Performance} {Embedded} {Architectures} and {Compilers}}}, Y.~N. Patt, P.~Foglia, E.~Duesterwald, P.~Faraboschi, and X.~Martorell, Eds.\hskip 1em plus 0.5em minus 0.4em\relax Berlin, Heidelberg: Springer, 2010, pp. 111--125.

\bibitem{liu_efficient_2013}
\BIBentryALTinterwordspacing
X.~Liu, M.~Smelyanskiy, E.~Chow, and P.~Dubey, ``Efficient sparse matrix-vector multiplication on x86-based many-core processors,'' in \emph{Proceedings of the 27th international {ACM} conference on {International} conference on supercomputing}, ser. {ICS} '13.\hskip 1em plus 0.5em minus 0.4em\relax New York, NY, USA: Association for Computing Machinery, Jun. 2013, pp. 273--282. [Online]. Available: \url{https://dl.acm.org/doi/10.1145/2464996.2465013}
\BIBentrySTDinterwordspacing

\bibitem{kreutzer_unified_2014}
\BIBentryALTinterwordspacing
M.~Kreutzer, G.~Hager, G.~Wellein, H.~Fehske, and A.~R. Bishop, ``A {Unified} {Sparse} {Matrix} {Data} {Format} for {Efficient} {General} {Sparse} {Matrix}-{Vector} {Multiplication} on {Modern} {Processors} with {Wide} {SIMD} {Units},'' \emph{SIAM Journal on Scientific Computing}, vol.~36, no.~5, pp. C401--C423, Jan. 2014, publisher: Society for Industrial and Applied Mathematics. [Online]. Available: \url{https://epubs.siam.org/doi/10.1137/130930352}
\BIBentrySTDinterwordspacing

\bibitem{zheng_biell_2014}
\BIBentryALTinterwordspacing
C.~Zheng, S.~Gu, T.-X. Gu, B.~Yang, and X.-P. Liu, ``{BiELL}: {A} bisection {ELLPACK}-based storage format for optimizing {SpMV} on {GPUs},'' \emph{Journal of Parallel and Distributed Computing}, vol.~74, no.~7, pp. 2639--2647, Jul. 2014. [Online]. Available: \url{https://www.sciencedirect.com/science/article/pii/S0743731514000458}
\BIBentrySTDinterwordspacing

\bibitem{ballard_hypergraph_2015}
\BIBentryALTinterwordspacing
G.~Ballard, A.~Druinsky, N.~Knight, and O.~Schwartz, ``Hypergraph {Partitioning} for {Parallel} {Sparse} {Matrix}-{Matrix} {Multiplication},'' in \emph{Proceedings of the 27th {ACM} symposium on {Parallelism} in {Algorithms} and {Architectures}}, ser. {SPAA} '15.\hskip 1em plus 0.5em minus 0.4em\relax New York, NY, USA: Association for Computing Machinery, Jun. 2015, pp. 86--88. [Online]. Available: \url{https://dl.acm.org/doi/10.1145/2755573.2755613}
\BIBentrySTDinterwordspacing

\bibitem{filippone_sparse_2017}
\BIBentryALTinterwordspacing
S.~Filippone, V.~Cardellini, D.~Barbieri, and A.~Fanfarillo, ``Sparse {Matrix}-{Vector} {Multiplication} on {GPGPUs},'' \emph{ACM Trans. Math. Softw.}, vol.~43, no.~4, pp. 30:1--30:49, Jan. 2017. [Online]. Available: \url{https://dl.acm.org/doi/10.1145/3017994}
\BIBentrySTDinterwordspacing

\bibitem{zhang_vectorized_2018}
\BIBentryALTinterwordspacing
H.~Zhang, R.~T. Mills, K.~Rupp, and B.~F. Smith, ``Vectorized {Parallel} {Sparse} {Matrix}-{Vector} {Multiplication} in {PETSc} {Using} {AVX}-512,'' in \emph{Proceedings of the 47th {International} {Conference} on {Parallel} {Processing}}, ser. {ICPP} '18.\hskip 1em plus 0.5em minus 0.4em\relax New York, NY, USA: Association for Computing Machinery, Aug. 2018, pp. 1--10. [Online]. Available: \url{https://dl.acm.org/doi/10.1145/3225058.3225100}
\BIBentrySTDinterwordspacing

\bibitem{pogorelov_performance_2023}
\BIBentryALTinterwordspacing
K.~Pogorelov, J.~Trotter, and J.~Langguth, ``Performance {Prediction} for {Sparse} {Matrix} {Vector} {Multiplication} {Using} {Structure}-{Dependent} {Features},'' in \emph{Euro-{Par} 2023: {Parallel} {Processing} {Workshops}: {Euro}-{Par} 2023 {International} {Workshops}, {Limassol}, {Cyprus}, {August} 28 – {September} 1, 2023, {Revised} {Selected} {Papers}, {Part} {I}}.\hskip 1em plus 0.5em minus 0.4em\relax Berlin, Heidelberg: Springer-Verlag, Aug. 2023, pp. 135--146. [Online]. Available: \url{https://doi.org/10.1007/978-3-031-50684-0_11}
\BIBentrySTDinterwordspacing

\bibitem{hwang_cerberus_2024}
\BIBentryALTinterwordspacing
S.~Hwang, D.~Baek, J.~Park, and J.~Huh, ``Cerberus: {Triple} {Mode} {Acceleration} of {Sparse} {Matrix} and {Vector} {Multiplication},'' \emph{ACM Trans. Archit. Code Optim.}, vol.~21, no.~2, pp. 38:1--38:24, May 2024. [Online]. Available: \url{https://dl.acm.org/doi/10.1145/3653020}
\BIBentrySTDinterwordspacing

\bibitem{Izbicki2020}
M.~Izbicki and C.~R. Shelton, ``Distributed {Learning} of {Non}-convex {Linear} {Models} with {One} {Round} of {Communication},'' in \emph{{ECML}-{PKDD}}, U.~Brefeld, E.~Fromont, A.~Hotho, A.~Knobbe, M.~Maathuis, and C.~Robardet, Eds.\hskip 1em plus 0.5em minus 0.4em\relax Cham: Springer International Publishing, 2020, pp. 197--212.

\bibitem{leblond17a}
\BIBentryALTinterwordspacing
R.~Leblond, F.~Pedregosa, and S.~Lacoste-Julien, ``{ASAGA}: {Asynchronous} {Parallel} {SAGA},'' in \emph{Proceedings of the 20th {International} {Conference} on {Artificial} {Intelligence} and {Statistics}}, A.~Singh and J.~Zhu, Eds., vol.~54.\hskip 1em plus 0.5em minus 0.4em\relax Fort Lauderdale, FL, USA: PMLR, 2017, pp. 46--54, series Title: Proceedings of Machine Learning Research. [Online]. Available: \url{http://proceedings.mlr.press/v54/leblond17a.html}
\BIBentrySTDinterwordspacing

\bibitem{Smith2016}
\BIBentryALTinterwordspacing
V.~Smith, S.~Forte, M.~I. Jordan, and M.~Jaggi, ``L1-{Regularized} {Distributed} {Optimization}: {A} {Communication}-{Efficient} {Primal}-{Dual} {Framework},'' in \emph{{ML} {Systems} {Workshop} at {International} {Conference} on {Machine} {Learning} ({ICML} '16)}, 2016, pp. 1--23, arXiv: 1512.04011. [Online]. Available: \url{http://arxiv.org/abs/1512.04011}
\BIBentrySTDinterwordspacing

\bibitem{10.1145/3637528.3672038}
\BIBentryALTinterwordspacing
F.~Lu, R.~R. Curtin, E.~Raff, F.~Ferraro, and J.~Holt, ``High-dimensional distributed sparse classification with scalable communication-efficient global updates,'' in \emph{Proceedings of the 30th ACM SIGKDD Conference on Knowledge Discovery and Data Mining}, ser. KDD '24.\hskip 1em plus 0.5em minus 0.4em\relax New York, NY, USA: Association for Computing Machinery, 2024, p. 2037–2047. [Online]. Available: \url{https://doi.org/10.1145/3637528.3672038}
\BIBentrySTDinterwordspacing

\bibitem{Langford2009}
\BIBentryALTinterwordspacing
J.~Langford, L.~Li, and T.~Zhang, ``Sparse online learning via truncated gradient,'' \emph{The Journal of Machine Learning Research}, vol.~10, pp. 777--801, 2009, arXiv: 0806.4686v2. [Online]. Available: \url{http://dl.acm.org/citation.cfm?id=1577097}
\BIBentrySTDinterwordspacing

\bibitem{Balakrishnan2008}
S.~Balakrishnan and D.~Madigan, ``Algorithms for {Sparse} {Linear} {Classifiers} in the {Massive} {Data} {Setting},'' \emph{Journal of Machine Learning Research}, vol.~9, pp. 313--337, 2008.

\end{thebibliography}

\clearpage
\appendix

\onecolumn

\section{Additional Results}

The full version of \autoref{tbl:main_results} is presented below, showing that standard Celer performs worse

\begin{table}[!h]
    \centering
\caption{Most papers consider only one, or no, datasets of these sizes. Yet these datasets are imminently reasonable to run on a modern desktop, and the minimum value of $C$ (right column) that produces $\mathbf{w}^*=0$ is always at least 5 orders of magnitude smaller than $C=1$, yet it is a common default. }
\label{tbl:dataset_sizes_full}
\adjustbox{max width=0.99\columnwidth}{%
\begin{tabular}{llrrrrrrr}
\hline
\multicolumn{2}{c}{Dataset}                                                          & \multicolumn{1}{c}{skglm}     & \multicolumn{1}{c}{Celer} & \multicolumn{1}{c}{Celer-PN}   & \multicolumn{1}{c}{SAGA}      & \multicolumn{1}{c}{Liblinear} & \multicolumn{1}{c}{OWL-QN}    & \multicolumn{1}{c}{LBFGS*} \\ \hline
                                                                              & f(w) & 0.0905                        & \textbf{0.0904}           & \textbf{0.0904}                & 0.0911                        & 0.0986                        & 0.1228                        & 0.1217                    \\
                                                                              & Acc  & 97.0                          & \textbf{97.1}             & \textbf{97.1}                  & 97.0                          & 96.8                          & 96.8                          & 96.8                      \\
                                                                              & nnz  & 79                            & 86                        & 87                             & 267                           & 155                           & 246                           & 515                       \\
\multirow{-4}{*}{\begin{tabular}[c]{@{}l@{}}URL\\ C=0.01\end{tabular}}        & Time & 33.2                          & 2010.7                    & 327.3                          & 219.9                         & \textbf{6.7}                  & 60.6                          & 84.5                      \\ \hline
                                                                              & f(w) & 0.0598                        &                           & {\color[HTML]{CB0000} 0.0381}  & \textbf{0.0396}               & 0.0571                        & 0.0740                        & 0.0528                    \\
                                                                              & Acc  & 97.9                          &                           & {\color[HTML]{CB0000} 98.7}    & \textbf{98.6}                 & 98.3                          & 97.5                          & 98.4                      \\
                                                                              & nnz  & 471                           &                           & {\color[HTML]{CB0000} 1681}    & 2391                          & 3056                          & 13925                         & 11843                     \\
\multirow{-4}{*}{\begin{tabular}[c]{@{}l@{}}URL\\ C=1\end{tabular}}           & Time & 31.5                          & \textgreater{}24 hours    & {\color[HTML]{CB0000} 8832.7}  & 1268.8                        & \textbf{8.7}                  & 42.4                          & 89.8                      \\ \hline
                                                                              & f(w) & \textbf{0.2224}               & 0.2225                    & 0.2225                         & 0.2227                        & 0.2227                        & 0.3050                        & 0.3095                    \\
                                                                              & Acc  & 92.5                          & 92.5                      & 92.5                           & 92.5                          & 92.5                          & 92.5                          & \textbf{92.7}             \\
                                                                              & nnz  & 48                            & 47                        & 47                             & 48                            & 50                            & 68                            & 177                       \\
\multirow{-4}{*}{\begin{tabular}[c]{@{}l@{}}Webspam\\ C=0.01\end{tabular}}    & Time & \textbf{18.8}                 & 179.1                     & 48.1                           & 520.4                         & 28.2                          & 559.5                         & 448.2                     \\ \hline
                                                                              & f(w) & 0.0417                        &                           &                                &                               & \textbf{0.0383}               & 0.0674                        & 0.0609                    \\
                                                                              & Acc  & 98.6                          &                           &                                &                               & \textbf{99.0}                 & 98.4                          & 98.8                      \\
                                                                              & nnz  & 486                           &                           &                                &                               & 1597                          & 6070                          & 6711                      \\
\multirow{-4}{*}{\begin{tabular}[c]{@{}l@{}}Webspam\\ C=1\end{tabular}}       & Time & 24.0                          & \textgreater{}24 hours    & \textgreater{}24 hours         & \textgreater{}24 hours        & \textbf{23.6}                 & 219.5                         & 310.4                     \\ \hline
                                                                              & f(w) & 0.3099                        &                           & \textbf{0.3093}                & {\color[HTML]{CB0000} 0.2989} & 0.3158                        & 0.3150                        & 0.3174                    \\
                                                                              & Acc  & 99.8                          &                           & 99.7                           & {\color[HTML]{CB0000} 99.2}   & 99.7                          & 99.7                          & \textbf{99.9}             \\
                                                                              & nnz  & 551                           &                           & 609                            & {\color[HTML]{CB0000} 10188}  & 663                           & 657                           & 1356                      \\
\multirow{-4}{*}{\begin{tabular}[c]{@{}l@{}}Avazu-app\\ C=0.01\end{tabular}}  & Time & 72.7                          & \textgreater{}24 hours    & 6236.1                         & {\color[HTML]{CB0000} 3042.1} & 35.7                          & 195.8                         & \textbf{30.0}             \\ \hline
                                                                              & f(w) & {\color[HTML]{CB0000} 0.3214} &                           & {\color[HTML]{CB0000} 0.2988}  & {\color[HTML]{CB0000} 0.2983} & \textbf{0.3006}               & {\color[HTML]{CB0000} 0.3769} & 0.3065                    \\
                                                                              & Acc  & {\color[HTML]{CB0000} 99.8}   &                           & {\color[HTML]{CB0000} 99.2}    & {\color[HTML]{CB0000} 99.2}   & 98.3                          & {\color[HTML]{CB0000} 84.0}   & \textbf{99.5}             \\
                                                                              & nnz  & {\color[HTML]{CB0000} 115}    &                           & {\color[HTML]{CB0000} 9597}    & {\color[HTML]{CB0000} 22928}  & 10209                         & {\color[HTML]{CB0000} 26}     & 10402                     \\
\multirow{-4}{*}{\begin{tabular}[c]{@{}l@{}}Avazu-app\\ C=1\end{tabular}}     & Time & {\color[HTML]{CB0000} 34.0}   & \textgreater{}24 hours    & {\color[HTML]{CB0000} 31768.6} & {\color[HTML]{CB0000} 9408.6} & \textbf{20.5}                 & {\color[HTML]{CB0000} 0.3}    & 47.1                      \\ \hline
                                                                              & f(w) & \textbf{0.4424}               &                           & {\color[HTML]{CB0000} 0.4416}  &                               & 0.4462                        & 0.4456                        & 0.4516                    \\
                                                                              & Acc  & 80.7                          &                           & {\color[HTML]{CB0000} 98.0}    &                               & 98.0                          & 98.0                          & \textbf{98.3}             \\
                                                                              & nnz  & 814                           &                           & {\color[HTML]{CB0000} 922}     &                               & 989                           & 1053                          & 3692                      \\
\multirow{-4}{*}{\begin{tabular}[c]{@{}l@{}}Avazu-site\\ C=0.01\end{tabular}} & Time & 148.8                         & \textgreater{}24 hours    & {\color[HTML]{CB0000} 16475.8} & \textgreater{}24 hours        & 75.1                          & 416.5                         & \textbf{34.8}             \\ \hline
                                                                              & f(w) & 0.4412                        &                           &                                &                               & \textbf{0.4369}               & 0.4367                        & 0.4446                    \\
                                                                              & Acc  & \textbf{98.0}                 &                           &                                &                               & 97.7                          & 97.7                          & \textbf{98.0}             \\
                                                                              & nnz  & 1007                          &                           &                                &                               & 11798                         & 12186                         & 11200                     \\
\multirow{-4}{*}{\begin{tabular}[c]{@{}l@{}}Avazu-site\\ C=1\end{tabular}}    & Time & 114.6                         & \textgreater{}24 hours    & \textgreater{}24 hours         & \textgreater{}24 hours        & \textbf{42.8}                 & 296.7                         & 54.9                      \\ \hline
                                                                              & f(w) & \textbf{0.3180}               &                           &                                &                               & 0.3326                        & 0.3324                        & 0.3347                    \\
                                                                              & Acc  & \textbf{88.8}                 &                           &                                &                               & 88.7                          & \textbf{88.8}                 & 88.5                      \\
                                                                              & nnz  & 3845                          &                           &                                &                               & 4002                          & 4394                          & 11130                     \\
\multirow{-4}{*}{\begin{tabular}[c]{@{}l@{}}kdda\\ C=0.01\end{tabular}}       & Time & 182.5                         & \textgreater{}24 hours    & \textgreater{}24 hours         & \textgreater{}24 hours        & \textbf{41.5}                 & 725.5                         & 71.5                      \\ \hline
                                                                              & f(w) & {\color[HTML]{CB0000} 0.3222} &                           &                                &                               & 0.2668                        & \textbf{0.2648}               & 0.2855                    \\
                                                                              & Acc  & {\color[HTML]{CB0000} 88.5}   &                           &                                &                               & 89.3                          & \textbf{90.4}                 & 89.5                      \\
                                                                              & nnz  & {\color[HTML]{CB0000} 3771}   &                           &                                &                               & 760537                        & 958627                        & 1576491                   \\
\multirow{-4}{*}{\begin{tabular}[c]{@{}l@{}}kdda\\ C=1\end{tabular}}          & Time & {\color[HTML]{CB0000} 59.8}   & \textgreater{}24 hours    & \textgreater{}24 hours         & \textgreater{}24 hours        & \textbf{65.8}                 & 869.7                         & 218.1                     \\ \hline
                                                                              & f(w) & \textbf{0.3143}               &                           &                                &                               & 0.3263                        & 0.3256                        & 0.3278                    \\
                                                                              & Acc  & \textbf{89.1}                 &                           &                                &                               & \textbf{89.1}                 & \textbf{89.1}                 & \textbf{89.1}             \\
                                                                              & nnz  & 8361                          &                           &                                &                               & 8509                          & 9533                          & 22932                     \\
\multirow{-4}{*}{\begin{tabular}[c]{@{}l@{}}kddb\\ C=0.01\end{tabular}}       & Time & 159.0                         & \textgreater{}24 hours    & \textgreater{}24 hours         & \textgreater{}24 hours        & \textbf{72.0}                 & 1048.4                        & 176.4                     \\ \hline
                                                                              & f(w) & {\color[HTML]{CB0000} 0.3291} &                           &                                &                               & 0.2523                        & \textbf{0.2520}               & 0.2814                    \\
                                                                              & Acc  & {\color[HTML]{CB0000} 88.9}   &                           &                                &                               & 89.8                          & \textbf{89.9}                 & 89.8                      \\
                                                                              & nnz  & {\color[HTML]{CB0000} 2711}   &                           &                                &                               & 1775533                       & 2202695                       & 2983383                   \\
\multirow{-4}{*}{\begin{tabular}[c]{@{}l@{}}kddb\\ C=1\end{tabular}}          & Time & {\color[HTML]{CB0000} 47.6}   & \textgreater{}24 hours    & \textgreater{}24 hours         & \textgreater{}24 hours        & \textbf{137.4}                & 688.9                         & 375.5                     \\ \hline
                                                                              & f(w) & 0.1714                        &                           &                                &                               & \textbf{0.1680}               & 0.1682                        & 0.1689                    \\
                                                                              & Acc  & 95.5                          &                           &                                &                               & \textbf{95.6}                 & \textbf{95.6}                 & \textbf{95.6}             \\
                                                                              & nnz  & 96                            &                           &                                &                               & 2654                          & 4499                          & 6050                      \\
\multirow{-4}{*}{\begin{tabular}[c]{@{}l@{}}kdd12\\ C=0.01\end{tabular}}      & Time & 132.1                         & \textgreater{}24 hours    & \textgreater{}24 hours         & \textgreater{}24 hours        & 309.4                         & 581.8                         & \textbf{280.4}            \\ \hline
                                                                              & f(w) & 0.1714                        &                           &                                &                               & \textbf{0.1573}               & 0.1612                        & 0.1663                    \\
                                                                              & Acc  & 95.5                          &                           &                                &                               & \textbf{95.6}                 & \textbf{95.6}                 & \textbf{95.6}             \\
                                                                              & nnz  & 103                           &                           &                                &                               & 342325                        & 635187                        & 406605                    \\
\multirow{-4}{*}{\begin{tabular}[c]{@{}l@{}}kdd12\\ C=1\end{tabular}}         & Time & 127.3                         & \textgreater{}24 hours    & \textgreater{}24 hours         & \textgreater{}24 hours        & 596.0                         & 639.9                         & \textbf{313.0}            \\ \hline
\end{tabular}
}
\end{table}

\clearpage

\section{Code}

\begin{lstlisting}[language=Python]
def trainLRviaLBFGS(lambda_penalty, X_train, y_train, X_test, y_test, line_search_steps = 30, max_steps=1000, ftol=1e-7, gtol=1e-5, logFile=None, reScaleInit=True, useBias=False, budget=5, min_steps=5, min_og_steps=5, use_hess_diag_est=False, nnz_rel_change=np.nan, modelSavePath=None, parallel=PARALLEL_RUN, dtype=None, w_start=None):
    n, d = X_train.shape

    if dtype is None:
        if w_start is not None:
            dtype = w_start.dtype
        else:
            dtype = X_train.dtype
    
    alpha = 1.0
    label_ratio = np.unique(y_train, return_counts=True)[1].min()/n
    
    f_prev = 0.0

    nnz_prev = 3*d
    y_train = y_train.astype(dtype)
    y_test = y_test.astype(dtype)

    # used for computing X.T @ w in parallel 
    if parallel:
        out_tmp = np.zeros((max_threads,d))
    else:
        out_tmp = None

    eps_out = 1.0
    
    start = time.time()
    if w_start is None: # make a guess 
        w_zero = np.zeros(d, dtype=dtype)
        #cacluate gradient from zero vector
        f_noPen, g_noPen, Xw = loss_grad_w_generic_no_penalty(lr_loss_fun, lr_grad_fun, w_zero, X_train, y_train, out_tmp=out_tmp, parallel=parallel)
        f_w0, g_  = loss_grad_w_generic(f_noPen, g_noPen, w_zero, lambda_penalty, alpha, X_train.shape, useBias=None)
        g_norm_1 = np.linalg.norm(g_,1)
        partial_function = functools.partial(logLoss_wrt_scale, y=y_train, penalty=lambda_penalty, Xw=X_train@(-g_), bias = 0.0, w_norm_1=g_norm_1)
        max_scale = 1.0
        best_rough = (0.0, f_w0)
        while max_scale < min(n, d)/2:
            f_s = partial_function(max_scale) 
            max_scale *= 2
            if f_s < best_rough[1]:
                best_rough = max_scale, f_s
            if f_s < f_w0:
                break

        loss_minimizing_scale = 1.0
        try:
            loss_minimizing_scale = scipy.optimize.brent(partial_function, brack=[0,max_scale], maxiter=1000, tol=1e-4)
            if loss_minimizing_scale <= 0:
                loss_minimizing_scale, _ = best_rough
        except: 
            loss_minimizing_scale = 1.0

        print(f"loss_minimizing_scale={loss_minimizing_scale}")
        # w_init = qq
        w_init = -g_
        w_init *= loss_minimizing_scale
        w_init = w_init.astype(dtype)
    else:
        w_init = w_start.astype(dtype)
    total_time_spent = time.time()-start

    x_k = w_init*1.0

    # useBias = True
    if useBias:
        x_k = np.hstack( ([0], x_k), dtype=dtype)
    else:
        x_k = x_k
    
    def f_and_g_nt(x):
        f_noPen, g_noPen, Xw = loss_grad_w_generic_no_penalty(lr_loss_fun, lr_grad_fun, x, X_train, y_train, out_tmp=out_tmp, parallel=parallel)
        f_, g_  = loss_grad_w_generic(f_noPen, g_noPen, x, lambda_penalty, alpha, X_train.shape, useBias=None)
        return f_, g_

    alphas = np.zeros((budget), dtype=dtype)
    S = np.zeros((budget, d), dtype=dtype)
    Y = np.ones((budget, d), dtype=dtype)
    inv_rhos = np.ones((budget), dtype=dtype)
    cur_budget_pos = -1 #this will be set to zero on the increment of the main loop
    budget_first_item = 0
    budget_last_item = 0
    budget_filled = 0

    

    f_prev = 0.0
    
    start = time.time()
    f_k, g_k = f_and_g_nt(x_k)
    g_k_inf = np.linalg.norm(g_k, np.inf)
    total_time_spent += time.time()-start

    ema_alpha = np.array(2.0/(budget*2+1), dtype=dtype) #make the ema_window 1x the budget window for the hessian
    H_k_ema_est = np.ones_like(x_k)


    if logFile is not None:
        outFile = open(f"{logFile}.csv", 'w')
        outFile.write("iteration, C, gtol, seconds, nnz, sparsity, train_acc, test_acc, norm_penality, train_loss, test_loss, f_tol, ftol_ema, active_set_size, spred\n")
        outFile.flush()

    active_dims = np.ones(d) > 0 #vector of all true
    X_active = X_train
    gtol_reclip = 0.0
    ftol_reclip = 0.01
    nnz_reclip = 0
    forceTransition = False
    previous_step_size = 1.0
   
    ftol_ema_log = 0.0
    ftol_ema_weight = 0.1
    
    for step in tqdm(range(max_steps)):
        if useBias:
            bias = x_k[0]
            if x_k.shape[0] <= X_active.shape[1]+1: #explicit w 
                v = x_k[1:]
                u = 1.0
            else: #reparameterization
                v = x_k[1:1+x_k.shape[0]//2]
                u = x_k[1+x_k.shape[0]//2:]
        else:
            bias = 0.0
            if x_k.shape[0] <= X_active.shape[1]+1: #explicit w 
                v = x_k
                u = 1.0
            else: #reparameterization
                v = x_k[:x_k.shape[0]//2]
                u = x_k[x_k.shape[0]//2:]
        w_candidate = v*u
        if active_dims.sum() < d: #expand this back out
            w_ = np.zeros(d, dtype=dtype)
            w_[active_dims] = w_candidate
            w_candidate = w_

        nnz = (np.abs(w_candidate) >= 1e-3).sum()
        sparsity = 1-nnz/w_candidate.shape[0]
        y_hat_train = mat_vec_csr(X_train, w_candidate) + bias
        y_hat_test = mat_vec_csr(X_test, w_candidate) + bias
        train_acc = (jnp.sign(y_hat_train) == y_train).sum()/y_train.shape[0]
        test_acc = (jnp.sign(y_hat_test) == y_test).sum()/y_test.shape[0]
        norm_penality = 1/(n*C_) * np.linalg.norm(w_candidate, 1) 
        train_loss = norm_penality + np.mean(log1pexp(-y_train * y_hat_train))
        test_loss = norm_penality + np.mean(log1pexp(-y_test * y_hat_test))

        if modelSavePath is not None:
            if step %
                np.save(f"{modelSavePath}_ceof_c_{C}_step_{step}.npy", w_candidate)

        g_k_inf = np.linalg.norm(g_k, np.inf)
        if g_k_inf <= 1e-8:
            break
        
        eps_ll = np.linalg.norm(g_k, 1)
        tol_ll = eps_ll/eps_out
        
        gtol_obv = g_k_inf/label_ratio
        ftol_obv = np.abs(f_k-f_prev)
        if ftol_ema_log == 0.0:
            ftol_ema_log = np.log(ftol_obv+1e-20)
        else:
            ftol_ema_log = ftol_ema_weight * np.log(ftol_obv+1e-20) + (1-ftol_ema_weight) * ftol_ema_log
        ftol_ema = np.exp(ftol_ema_log)
        
        print(f"{step}, {f_k}, {previous_step_size:.3f}, {total_time_spent:.2f}, {nnz}, {ftol_obv}, {ftol_ema}, {gtol_obv}, {tol_ll}, {test_acc:.5f}")
        if logFile is not None:
            outFile.write(f"{step}, {C_:.4f}, {gtol_obv}, {total_time_spent},{nnz}, {sparsity}, {train_acc}, {test_acc}, {norm_penality}, {train_loss}, {test_loss}, {ftol_obv}, {ftol_ema}, {active_dims.sum()}, {spred}\n")
            outFile.flush()
            os.fsync(outFile.fileno())

        if g_k_inf <= gtol*label_ratio and step > min_steps:
            print(f"Exit for gtol {g_k_inf} <= {gtol*label_ratio}")
            break
        if ftol_ema <= ftol and step > min_steps:
            print(f"Exit for ftol {ftol_ema} <= {ftol*label_ratio}")
            break

        start = time.time()
        projection_step_bc_no_og_progress = x_k.shape[0] == d and relChange(nnz_prev, nnz) < nnz_rel_change
        if step < min_og_steps:
            projection_step_bc_no_og_progress = False
        projection_step_bc_ftol_prog = (ftol_ema < ftol_reclip)
        
        if projection_step_bc_no_og_progress or projection_step_bc_ftol_prog:
            #lets clip things we know are non-viable toward zero
            x_k = w_candidate
            f_noPen, g_noPen, Xw = loss_grad_w_generic_no_penalty(lr_loss_fun, lr_grad_fun, x_k, X_train, y_train)
            f_k, g_k = loss_grad_w_generic(f_noPen, np.copy(g_noPen), x_k, lambda_penalty, alpha, X_train.shape)
            #at this point, we know shape(x_k) = d, and x_k is in the orignal space
            s = C_*n #scale to apply
            # L_p = g_k*s - alpha*np.sign(x_k)
            L_p = g_noPen*s
            v_j_max = np.where(x_k == 0, np.maximum(L_p-alpha, -alpha-L_p), np.where(x_k > 0, np.abs(L_p+alpha), np.abs(L_p-alpha) )).max()/n
            min_active = min(np.abs(L_p[np.abs(x_k) > 1e-3]).min()*1,1.0)
            projection_clip = 1.0 #max((alpha-v_j_max)*0.1, min_active)
            for clip_line_search_trial in range(20):
                active_dims = (np.abs(L_p) >= projection_clip)
                x_trial = np.where(active_dims, x_k, 0.0)
                f_noPen_trial, g_noPen_trial, Xw = loss_grad_w_generic_no_penalty(lr_loss_fun, lr_grad_fun, x_trial, X_train, y_train)
                f_trial, g_trial = loss_grad_w_generic(f_noPen_trial, g_noPen_trial, x_trial, lambda_penalty, alpha, X_train.shape) #TODO, just evalu f_trail for faster speed
                if f_trial <= f_k *1.001:
                    break
                projection_clip /= 2
            x_k = x_k[active_dims]
            X_active = X_train[:,active_dims]
            print(f"\tActive reduction to {active_dims.sum()} but max active is {min_active}")
            nnz_reclip = (np.abs(x_k) > 1e-3).sum()/2.0 #if we reduce nnz by a factor of 10, worth re-clipping

            if parallel: #re-size the temp space to the correct dimension so that we don't get dimension miss-match issues 
                out_tmp = np.zeros((max_threads,X_active.shape[1]))
            
            def f_and_g_nt(x):
                f_noPen, g_noPen, Xw = loss_grad_w_generic_no_penalty(lr_loss_fun, lr_grad_fun, x, X_active, y_train, out_tmp=out_tmp, parallel=parallel)
                f_, g_  = loss_grad_w_generic(f_noPen, g_noPen, x, lambda_penalty, alpha, X_active.shape, useBias=None)
                return f_, g_

            S = np.zeros((budget, x_k.shape[0]), dtype=dtype)
            Y = np.ones((budget, x_k.shape[0]), dtype=dtype)
            budget_first_item, budget_last_item, budget_filled = 0, 0, 0
            f_k, g_k = f_and_g_nt(x_k)
            gtol_reclip = gtol_obv/100.0
            ftol_reclip = ftol_ema/10.0

        H_k_0 = None
        # p_k = -H_k grad(f_k); (6.18)
        p_k = -twoLoopHp_parallel(g_k, budget_first_item, budget_filled, inv_rhos, S, Y, alphas, H_k_0=H_k_0)
        # p_k = -g_k
        # Set x_k+1 = x_k + a_k p_k where a_k comes from a line search
        a, f_kp1, _, g_kp1, x_kp1 = armijo_line_search(f_and_g_nt, x_k, f_k, g_k, p_k, alpha_0=1.0, max_steps=5)
        
        if a <= 1e-6: 
            #lets try armijo w/o two-loop first
            print(f"\tBad armijo {a}, trying w/o history")
            a, f_kp1, _, g_kp1, x_kp1 = armijo_line_search(f_and_g_nt, x_k, f_k, g_k, -g_k, alpha_0=1.0, max_steps=5)
            if a <= 1e-2: #Ok, we really need a HZ line search
                a, f_kp1, _, g_kp1 = hz_lineSearch(f_and_g_nt, step, x_k, f_k, g_k, p_k, previous_step_size, gamma=0.66, max_line_steps = line_search_steps)
                x_kp1 = x_k + a * p_k
                print(f"hz picked a={a} and budget is {budget_filled}")
                if a == 0 and budget_filled == 0:
                    if step < min_steps:
                        a = 1e-3
                        x_kp1 = x_k + a * p_k
                        f_kp1, g_kp1 = f_and_g_nt(x_kp1)
                    else:
                        break
            else: #no history worked, clear it!
                print(f"armijo picked a={a} and budget is {budget_filled}")
                S = np.zeros((budget, x_k.shape[0]), dtype=dtype)
                Y = np.ones((budget, x_k.shape[0]), dtype=dtype)
                budget_first_item, budget_last_item, budget_filled = 0, 0, 0
        
        if a <= 1e-8:
            if nnz_rel_change > 0.0:
                forceTransition = True
                continue
            else:
                if budget_filled > 1:
                    S = np.zeros((budget, x_k.shape[0]), dtype=dtype)
                    Y = np.ones((budget, x_k.shape[0]), dtype=dtype)
                    budget_first_item, budget_last_item, budget_filled = 0, 0, 0
                    continue
                else:
                    print(f"Exit for tiny step size step size {a}")
                    break
        
        previous_step_size = a
        # x_kp1 = x_k + a * p_k
        # f_kp1, g_kp1 = f_and_g_nt(x_kp1)
        # Define s_k = x_k+1 - x_k and y_k = grad(f_k+1) - grad(f_k);
        # b/c (x_kp1 - x_k) = (x_k + a * p_k) - x_k = a*p_k
        S[budget_last_item,:] = p_k
        S[budget_last_item,:] *= a
        Y[budget_last_item,:] = g_kp1
        Y[budget_last_item,:] -= g_k
        inv_rhos[budget_last_item] = S[budget_last_item,:].dot(Y[budget_last_item,:])
        budget_last_item = (budget_last_item + 1 ) %
        if budget_filled >= budget:
            budget_first_item = (budget_first_item + 1 ) %
        else:
            budget_filled += 1
        end = time.time()
        total_time_spent += (end-start)
        f_prev = f_k*1.0
        x_k, f_k, g_k = x_kp1*1.0, f_kp1*1.0, g_kp1*1.0
        nnz_prev = nnz

    bias = 0.0
    if modelSavePath is not None:
        np.save(f"{modelSavePath}_ceof_c_{C}_step_{step}.npy", w_candidate)
    return w_candidate, bias
\end{lstlisting}

\end{document}